\documentclass[10pt,journal,compsoc]{IEEEtran}

\usepackage{cite}
\usepackage{url}
\usepackage{ragged2e}
\usepackage{epsfig}
\usepackage{graphicx}
\usepackage{amsmath,amssymb} 
\usepackage{subfigure}
\usepackage{algorithm}
\usepackage{algorithmicx}
\usepackage{algpseudocode}
\usepackage{threeparttable}
\usepackage{color}
\usepackage[normalem]{ulem}
\usepackage{multirow}
\usepackage{float}
\usepackage{amsfonts}
\usepackage{bm}
\usepackage{array}
\usepackage[table]{xcolor}
\usepackage{colortbl}
\usepackage{pifont}
\usepackage{diagbox}
\usepackage{rotating}
\usepackage{booktabs}
\usepackage{overpic}
\usepackage{textcomp}
\usepackage{contour}
\usepackage{enumitem}

\usepackage{algorithm}
\usepackage{algpseudocode}
\usepackage{amsmath}

\definecolor{citeGreen}{RGB}{69, 137, 51}
\usepackage[colorlinks=true,linkcolor=black,citecolor=citeGreen]{hyperref}

\definecolor{mygray}{gray}{.90}
\def\ie{\textit{i.e.}}
\def\eg{\textit{e.g.}}

\def\etal{{\textit{et al.~}}}

\newcommand{\figref}[1]{Fig.~\ref{#1}}
\newcommand{\tabref}[1]{Table~\ref{#1}}

\newcommand{\secref}[1]{Section \ref{#1}}

\definecolor{deeppink}{RGB}{220,20,60}
\definecolor{navyblue}{RGB}{0,191,255}
\definecolor{myblue}{rgb}{0.5, 0.8, 0.9}
\definecolor{myred}{rgb}{0.9, 0.2, 0.1}
\definecolor{mygreen}{rgb}{0.1, 0.8, 0.1}
\definecolor{mypurple}{rgb}{0.5, 0.2, 0.9}
\definecolor{mygrey}{rgb}{0.7, 0.7, 0.7}

\definecolor{egnet_blue}{rgb}{0.1, 0.5, 1}
\definecolor{minet_yellow}{rgb}{0.8, 0.6, 0.1}

\newcommand{\supp}[1]{\textcolor{magenta}{#1}}

\def\ourmodel{ICON}
\graphicspath{{./Imgs/}}
\newcommand{\tabincell}[2]{\begin{tabular}{@{}#1@{}}#2\end{tabular}}

\newcommand{\Rev}[1]{{#1}}

\newcommand{\Red}[1]{\textcolor{black}{#1}}

\begin{document}
%
\title{Salient Object Detection via Integrity Learning}
%
%
%
%

\author{
Mingchen Zhuge, 
Deng-Ping Fan$^{\dagger}$, 
Nian Liu,
Dingwen Zhang,~
Dong Xu,~\IEEEmembership{Fellow,~IEEE}, \\
and Ling Shao,~\IEEEmembership{Fellow,~IEEE} 
\IEEEcompsocitemizethanks{
\IEEEcompsocthanksitem The first two authors share equal contributions. 

\IEEEcompsocthanksitem Deng-Ping Fan is with the College of Computer Science, Nankai University, Tianjin, China. (Email: dengpfan@gmail.com)
\IEEEcompsocthanksitem Mingchen Zhuge and Nian Liu are with the IIAI, Abu Dhabi, UAE. 
(Email: mczhuge@gmail.com, liunian228@gmail.com)
\IEEEcompsocthanksitem Dingwen Zhang is with the Brain and Artificial Intelligence Laboratory, School of Automation, Northwestern Polytechnical University, Xi’an 710072, China. (Email: zhangdingwen2006yyy@gmail.com)
\IEEEcompsocthanksitem Dong Xu is with Dong Xu is with Department of Computer Science, The University of Hong Kong, Pokfulam, Hong Kong (Email: dongxudongxu@gmail.com)
\IEEEcompsocthanksitem Ling Shao is with Terminus Group, Beijing, China (Email: ling.shao@ieee.org).
\IEEEcompsocthanksitem $\dagger$ The work was done while Mingchen Zhuge was an intern at IIAI mentored by Deng-Ping Fan. Corresponding author: Nian Liu and Dingwen Zhang.
}
}

%
%

\markboth{IEEE Transactions on Pattern Analysis and Machine Intelligence}%
{Shell \MakeLowercase{\textit{et al.}}: Bare Demo of IEEEtran.cls for Computer Society Journals}
%



\IEEEtitleabstractindextext{%
\begin{abstract}
\justifying
	Although current salient object detection (SOD) works have achieved significant progress, they are limited when it comes to the integrity of the predicted salient regions.
	We define the concept of integrity at both a micro and macro level. Specifically, at the micro level, the model should highlight all parts that belong to a certain salient object. Meanwhile, at the macro level, the model needs to discover all salient objects in a given image.
	To facilitate integrity learning for SOD, we design a novel \textbf{I}ntegrity \textbf{Co}gnition \textbf{N}etwork (\textbf{\ourmodel}), which explores three important components for learning strong integrity features.
	1) Unlike existing models, which focus more on feature discriminability, we introduce a diverse feature aggregation (DFA) component to aggregate features with various receptive fields (\ie,~kernel shape and context) and increase feature diversity. Such diversity is the foundation for mining the integral salient objects. 
	2) Based on the DFA features, we introduce an integrity channel enhancement (ICE) component with the goal of enhancing feature channels that highlight the integral salient objects, 
    while suppressing the other distracting ones.
    3) After extracting the enhanced features, the part-whole verification (PWV) method is employed to determine whether the part and whole object features have strong agreement. Such part-whole agreements can further improve the micro-level integrity for each salient object.
    \Rev{To demonstrate the effectiveness of our ICON, comprehensive experiments are conducted on seven challenging benchmarks. Our ICON outperforms the baseline methods in terms of a wide range of metrics. Notably, our ICON achieves $\sim$10\% relative improvement over the previous best model in terms of average false negative ratio (FNR), on six datasets.}
	Codes and results are available at: 
	\supp{\href{https://github.com/mczhuge/ICON}{https://github.com/mczhuge/ICON}}.
\end{abstract}

}

\maketitle
\IEEEdisplaynontitleabstractindextext

%
\IEEEpeerreviewmaketitle


\IEEEraisesectionheading{\section{Introduction}}

\IEEEPARstart{S}{alient} object detection (SOD) aims to imitate the human visual perception system to capture the most significant regions in a given image~\cite{borji2015salient,fan2019rethinking,wang2019salient1}. As SOD is widely used in the field of computer vision, it plays a vital role in many downstream tasks, such as object detection \cite{zhang2019IJCV}, image retrieval~\cite{liu2013model}, co-salient object detection~\cite{deng2020re}, multi-modal matching~\cite{Zhuge2021KaleidoBERT}, VR/AR applications~\cite{qin2020u2} and semantic segmentation \cite{hoyer2019grid,wei2016stc,zeng2019joint}. 

Traditional SOD methods \cite{borji2015salient,wang2017saliency} predict saliency maps in a bottom-up manner, and are mainly based on handcrafted features, such as color contrast \cite{cheng2014global,itti1998model}, boundary backgrounds \cite{yang2013saliency,zhu2014saliency}, or center priors \cite{jiang2013salient}. To improve the representation capacity of the features used in SOD, current models employ convolutional neural network (CNN) or fully convolutional network architectures, which enable powerful feature learning processes to replace manually designed features. These methods have achieved remarkable progress and pushed the performance of SOD to a new level. More details of recent deep learning based SOD methods can be found in the surveys/benchmarks \cite{fan2018SOC,borji2014salient,wang2019salient1,borji2015salient,han2018advanced}.

\begin{figure}[t!]
	\centering
	\begin{overpic}[width=0.45\textwidth]{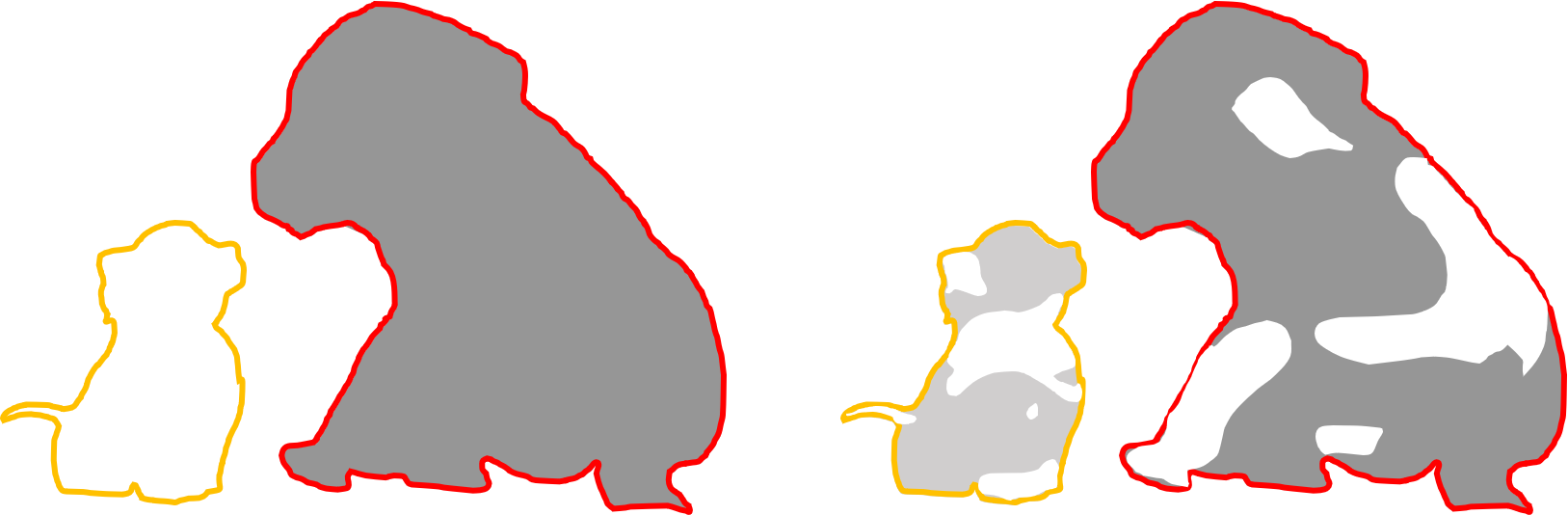} 
    \put(10.5,-3.1){\small{(a) macro integrity}}
    \put(64.5,-3.1){\small{(b) micro integrity}}  
	\end{overpic}
    \caption{
    \Rev{Integrity issues (\ie, (a) macro integrity and (b) micro integrity) from SOD. 
    The red and yellow contours represent the ground-truths. Grey regions denote the prediction results. 
    The drawing style was inspired by~\cite{cheng2021boundary}.
    }
    }\label{fig:first}
    \vspace{-8pt}
\end{figure} 

%

    
	

The current success in building deep learning based salient object detectors is mainly due to the use of \textit{multi-scale/level feature aggregation}, \textit{contextual modeling}, \textit{top-down modeling}, and \textit{edge-guided learning} mechanisms. 
Specifically, models with a multi-scale/level feature aggregation mechanism enhance the features from different levels and scales of the network, and then fuse them to generate the final SOD results. These approaches can help discover salient objects of various sizes and highlight the salient regions under the guidance of both coarse semantics and fine details. For example, the network proposed by Zhang \etal\cite{zhang2017amulet} first adaptively fuses multi-level features at five different scales, and then use them to generate predictions. Similarly, Luo \etal\cite{8100181} proposed to extract the global and local features at the low and high feature scales, respectively, and then fuse them to generate the final \Rev{results}. 

\begin{figure*}[t!]
	\centering
	\begin{overpic}[width=\textwidth]{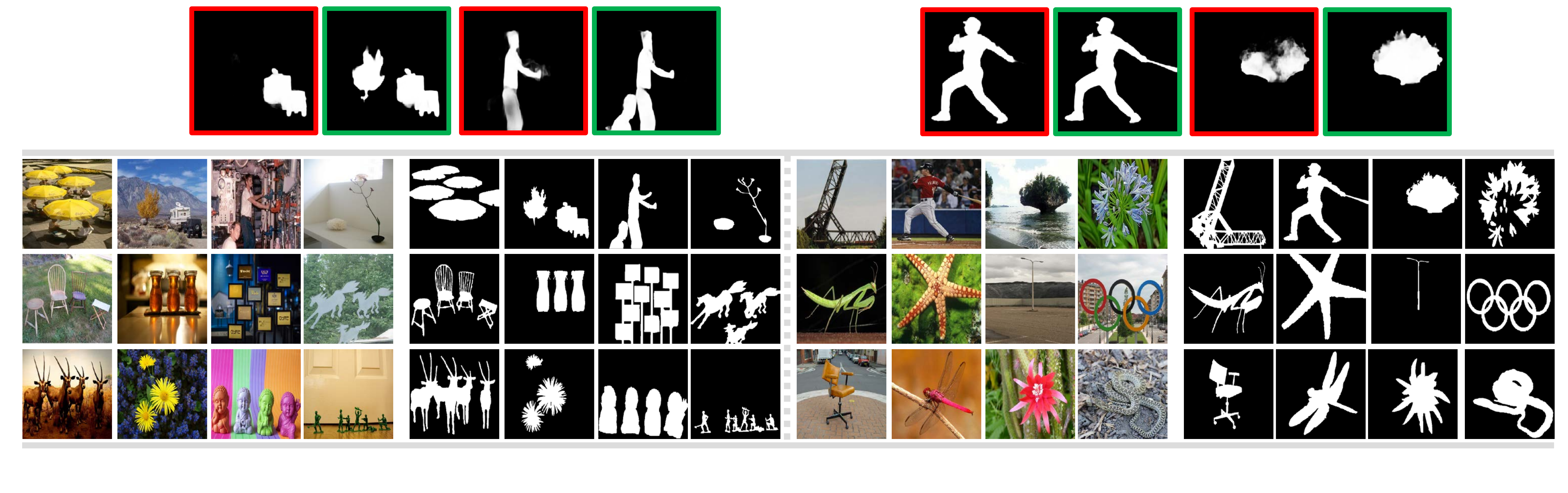} 
    \put(18.5,1.1){\small{(a) macro integrity}}
    \put(67.5,1.1){\small{(b) micro integrity}}  \put(2.5,26.4){\small{prediction}}  
    \put(49.5,26.4){\small{prediction}} 
	\end{overpic}
	\vspace{-23pt}
	\caption{\Red{Integrity is a good indicator for saliency prediction. } \Red{Here are some samples from the SOD datasets, with images (left) and ground truth (right) listed. To better predict the group of images, \eg, (a) the model needs to have the judgment of macro-level integrity, while the model needs to extract micro-level integrity efficiently present
	second group of images (\ie, (b)).} 
	\Red{On the top, several predictions with different integrity qualities are presented.} 
	Note that:~ \textcolor{myred}{$\blacksquare$}~Bad predictions,  ~\textcolor{green}{$\blacksquare$}~Comparably high-quality prediction results.}
	\label{fig:goodornot}
\end{figure*}
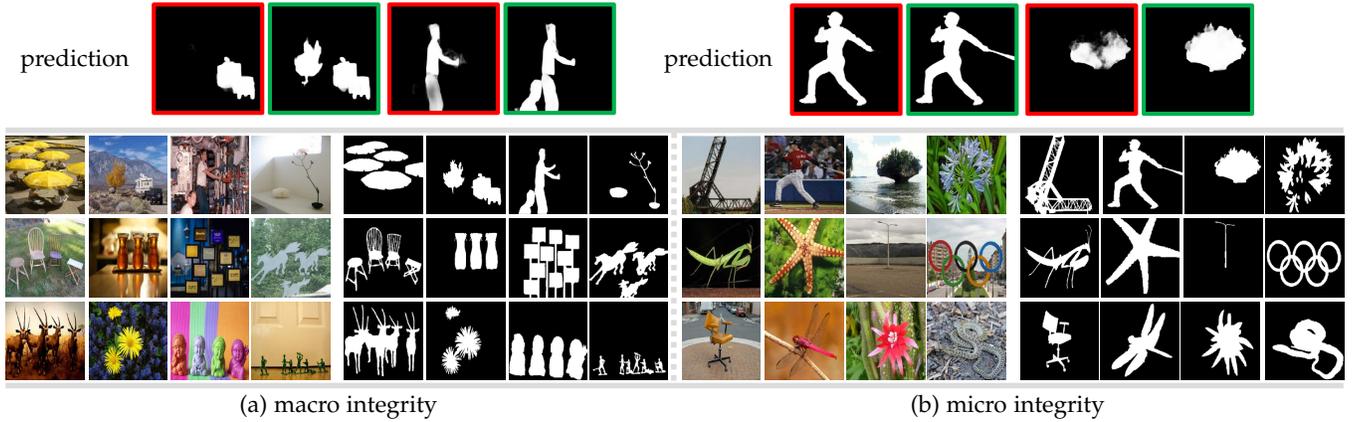 

Contextual modeling is another key mechanism in \Rev{SOD}. It helps infer the saliency of each local region by considering the surrounding contextual information. Current studies in the field of SOD usually design various attention modules to explore such information. Specifically, Zhao \etal\cite{zhao2019pyramid} proposed a pyramid feature attention network, where channel attention and spatial attention modules are introduced to process high- and low-level features, respectively, and consider the contextual information in different feature channels and spatial locations. Liu \etal\cite{liu2020picanet} proposed to learn a pixel-wise contextual attention for \Rev{SOD}. Deep models learned with such an attention module can infer the relevant importance between each pixel and its global/local context location, and thus achieve the selective aggregation of contextual information.

For top-down modeling, some SOD methods adopt carefully designed decoders to gradually infer salient regions under the guidance of high-level semantic cues. For example, Wang \etal\cite{wang2019iterative} built an iterative and cooperative inference network for \Rev{SOD}, where multiple top-down network streams work together with the bottom-up network streams in an iterative inference manner. Zhao \etal\cite{zhao2020suppress} proposed a gated dual-branch decoding structure to achieve cooperation among different levels of features in the top-down flow, which improves the discriminability of the whole network. In \cite{liu2019simple}, Liu \etal adopted a pyramid pooling to build global guiding features, to improve the top-down flow modeling.  

In order to accurately predict salient object boundaries, another group of methods introduce additional network streams or learned objective functions to force the network to pay more attention to the contours that separate the salient objects from the surrounding background. For example, Wei \etal\cite{wei2020label} built a label decoupling framework for SOD, which explicitly decomposes the original saliency map into a body map and a details map. Specifically, the body map concentrates on the central areas of the salient objects, while the details map focuses on the regions around the object boundaries.  
To improve the prediction precision of the salient contours and reduce the local noise in salient edge predictions, Wu \etal\cite{wu2019mutual}
proposed the mutual learning strategies to separately guide the foreground contour and edge detection tasks. 

Although the aforementioned mechanisms can improve the \Rev{SOD} performance in several aspects, the detection results produced are still not optimal. In our opinion, this is likely due to the under-exploration of another helpful and important mechanism, \ie,~the \textit{integrity learning} mechanism \Rev{(see \figref{fig:first}~(a) \& (b))}. \Rev{In this work}, we define the integrity  learning mechanism at two levels. At the micro level, the model should focus on part-whole relevance within a single salient object. At the macro level, the model needs to identify all salient objects within the given image scene. \Rev{In \figref{fig:goodornot}, we \Rev{present} some examples of the integrity qualities at both the macro and micro levels. It is clear that there exists a strong correlation between integrity and prediction performance.}

In order to pursue two-level integrity, we introduce three key components in our deep neural network design. The first is diverse feature aggregation (DFA). Unlike the existing models, which focus more on feature discriminability, DFA aggregates the features from various receptive fields (in terms of both the kernel shape and context) to increase their diversity. Such feature diversity provides the foundation for mining integral salient objects, since it considers richer contextual patterns to determine the activation of each neuron. The second component is called integrity channel enhancement (ICE), which aims at enhancing the feature channels that highlight the integral salient objects (at both the micro and macro levels), while suppressing the other distracting ones. As it is rare for the feature channels enhanced by ICE to perfectly match the real salient object regions, we further adopt a part-whole verification (PWV) component to judge whether the part features and whole features have a strong agreement to form the integral objects. This can help further improve integral learning at the micro level.

It is worth mentioning that some existing works have also tried to solve the macro-level integrity issue by introducing the auxiliary task for learning deep salient object detectors \cite{amirul2018revisiting,8237382}. However, these methods require additional supervision information on the number of salient objects within each image. In contrast, our newly proposed approach can tackle both macro- and micro-level integrity issues within a unified and entirely different learning framework, without requiring any additional supervision. 

Our overall framework for integrity learning is called the Integrity Cognition Network (\ourmodel), details of which are shown in \figref{fig:model}. Specifically, our \ourmodel~first leverages five convolutional blocks for basic feature extraction. Then, it passes the deep features at each level to a diverse feature aggregation module to extract the different feature bases. Next, the diverse feature bases extracted from three adjacent feature levels are sent to an integrity channel enhancement module. Here, an integrity guiding map is generated and then used to guide the attention weighting of each feature channel. Finally, the integrity channel enhancing features produced from the \Rev{three} 
feature levels are combined and passed through the part-whole verification module, which is implemented by using the capsule routing layers~\cite{hinton2018matrix}. After further verifying the agreement between the object parts and whole regions, the missing parts will be reinforced.
To sum up, this work includes three main contributions:

\begin{itemize}
	\item We investigate the integrity issue in SOD, which is essential yet under-studied in this field. 
	\vspace{5pt}
	\item We introduce three key components for achieving integral SOD, namely diverse feature aggregation, integrity channel enhancement, and part-whole verification. 
	\vspace{5pt}
	\item We design a novel network, \ie,~\ourmodel, that incorporates the three components and demonstrate its effectiveness on seven challenging datasets. In addition to its prominent performance, our approach also achieves real-time speed (\Rev{$\sim$60fps}).
\end{itemize}

The remainder of the paper is organized as follows. In \secref{sec2}, we discuss the related works. 
Then, we describe the proposed \ourmodel~in detail (see \secref{sec3}). Experimental results, including performance evaluations and comparison, are provided in \secref{sec4}. Finally, conclusions are drawn in \secref{sec5}.


\section{Related Work}\label{sec2}

\Rev{Over the past several decades, a number of SOD methods have been proposed and have achieved encouraging performance on various benchmark datasets. These existing SOD methods can be roughly categorized into scale learning based, boundary learning based, and integrity learning based approaches.}

\subsection{Scale Learning Approaches for SOD}
\Rev{Scale variation is one of the major challenges for SOD. Many works have tried to handle this issue from different perspectives.
%
%
Inspired by the HED model~\cite{xie2017holistically} for edge detection, 
%
%
DSS~\cite{hou2019deeply} introduced deep-to-shallow side-outputs with rich semantic features. This design enables shallow layers to distinguish real salient objects from the background, while retaining high resolution.
\Rev{In addition,} Zhang~\etal\cite{zhang2017amulet} designed a multi-level feature aggregation framework and employed the hierarchical features as the saliency cues for final saliency prediction. 
%
Meanwhile, RADF~\cite{hu18recurrently} integrated multi-level features and refines them within each layer with a recurrent pattern, which effectively suppresses the non-salient noise in lower layers and increases the salient details of features in higher layers.
Further, Zhao~\etal\cite{zhao2019optimizing} proposed to use the F-measure loss, which can generate precise contrastive maps to help segment multi-scale objects.
To efficiently extract multi-scale features, Pang~\etal\cite{pang2020multi} embedded self-interaction modules into their decoder units to learn the integrated information.
%
In the more recent work, GateNet~\cite{GateNet} adopted Fold-ASPP to gather multi-scale saliency cues.
Finally, Liu~\etal\cite{liu2021rethinking} utilized a centralized information interaction strategy to simultaneously process multi-scale 
features. 
%
}

\subsection{Boundary Learning Approaches for SOD}
\Rev{Boundary learning plays another important role for improving \Rev{SOD} results. Early works used boundary learning via biologically inspired methods~\cite{itti1998model, ma2003contrast, harel2006graph}. However, these models exhibit undesirable blurring results and usually lose entire salient areas. 
The more recent CNN-based approaches, which operate at the patch level (instead of pixel level), also suffer from blurred edges, due to the stride and pooling operations.
To address this issue, several works (\eg, \cite{hu2017deep}) use the pre-processing technology (\eg, superpixel~\cite{he2015supercnn}) to preserve the object boundaries, while other works, such as DSS~\cite{hou2019deeply}, DCL~\cite{li2016deep}, and PiCANet~\cite{liu2018picanet}, employed post-processing (\eg, conditional random fields~\cite{lafferty2001conditional}) to enhance edge details. The main drawback of these approaches is their slow inference speed. 
%
To learn the intrinsic edge information, PoolNet~\cite{liu2019simple} employed an auxiliary module for edge detection. 
Besides, many other works have improved edge quality by introducing boundary-aware loss functions. For instance, the recent works~\cite{luo2017non,li2018contour, su2019selectivity, zhao2019egnet, zhao2019pyramid} used explicit boundary losses to guide the learning of boundary details. 
%
Considering that the cross entropy loss prefers to predict hard pixel samples (\eg, 0 or 1) as non-integer values, BASNet~\cite{qin2019basnet} introduced a new prediction-refinement network and hybrid loss.
Dealing with the inherent defect of blurry boundaries, HRSOD~\cite{zeng2019towards} introduced the first high-resolution SOD dataset, which explores how high-resolution data can improve the performance of the salient object edges.
%
F3Net~\cite{F3Net} demonstrated that assigning larger weights to boundary pixels in the loss functions is a simple way to handle boundary problems.
%
In addition, the recent works such as SCRN~\cite{wu2019stacked}, LDF~\cite{wei2020label}, VST~\cite{liu2021visual} built two-stream architectures to model salient objects and boundaries simultaneously.
}

\begin{figure*}[t!]
	\centering
	\includegraphics[width=.98\textwidth]{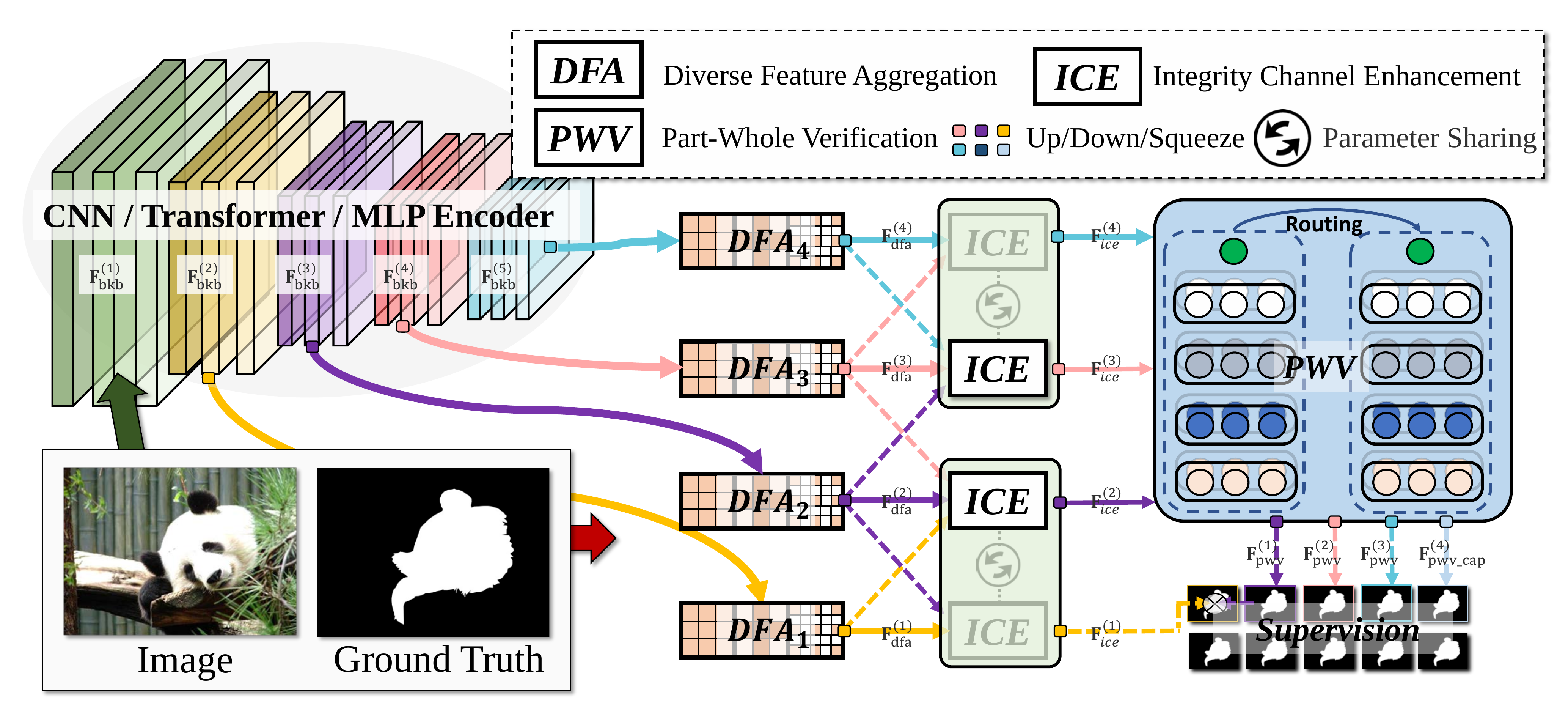} 
	\caption{Overall architecture of the proposed \ourmodel. Feature extraction block: \Rev{$\mathbf{F}^{(1)}_{bkb}$ - $\mathbf{F}^{(5)}_{bkb}$} denote different layers from ResNet-50~\cite{he2016deep}. Component 1: the diverse feature aggregation (DFA) module aggregates the features with various receptive fields. Component 2: the integrity enhancement (ICE) module aims at enhancing the feature channels that highlight the potential integral salient object. Component 3: the part-whole verification (PWV) module judges whether the part and whole features have strong agreement. 
	}
	\label{fig:model}
\end{figure*}

\subsection{Integrity Learning Approaches for SOD}
\Rev{Integrity learning 
is an under-explored research topic in SOD. 
%
Among the limited existing models, DCL~\cite{li2016deep} processed contrast information at both the pixel and patch levels in order to  simultaneously integrate global and local structural information.
CPD~\cite{wu2019cascaded} utilized an effective decoder to summarize the discriminative features, and segment the integral salient objects with the aid of holistic attention modules.
TSPOANet~\cite{liu2019employing} modeled part-object relationships in SOD, and produced better wholeness and uniformity scores for segmented salient objects with the help of a capsule network.
%
GCPANet~\cite{GCPANet} made full use of global context to capture the relationships between multiple salient objects or regions, and alleviate the dilution effect of features.
Wu~\etal\cite{wu2020deeper} used a bi-stream
network combining two feature backbones and gate control units to fuse complementary information.
Recently, transformers have become a hot research area in the field of computer vision. Mao~\etal\cite{mao2021transformer} proposed a transformer-based architecture for the context learning problem, which can also be considered as an integrity learning based approach.
%
%
}

\section{Framework}\label{sec3}
\subsection{Overview of \ourmodel}\label{Overview}
As shown in \figref{fig:model}, our method is based on an encoder-decoder architecture. The encoder uses ResNet-50 as the backbone to extract multi-level features. Meanwhile, the decoder integrates these multi-level features and generates the saliency map with multi-layer supervision.
For simplicity, from then on we denote the features generated by the backbone as a set  \Red{$\mathcal{F}_{bkb}=\{\mathbf{F}^{(0)}_{bkb}, \mathbf{F}^{(1)}_{bkb}, \mathbf{F}^{(2)}_{bkb}, \mathbf{F}^{(3)}_{bkb}, \mathbf{F}^{(4)}_{bkb}\}$}. To improve the computational efficiency, we do not use \Red{$\mathbf{F}^{(0)}_{bkb}$} in the decoder due to its large spatial size.

Next, we enhance the backbone features by passing them through the diverse feature aggregation (DFA) module, which consists of various convolutional blocks. Thereafter, we further use the integrity channel enhancement (ICE) module to strengthen the responses of the integrity-related channels
and coarsely highlight the integral salient parts. Finally, to further refine the saliency map, we utilize the part-whole verification (PWV) module to verify the agreement between object parts and the whole salient region.

\subsection{Diverse Feature Aggregation}\label{DFA}

\begin{figure*}[t!]
   \centering
	\begin{overpic}[width=0.85 \textwidth]{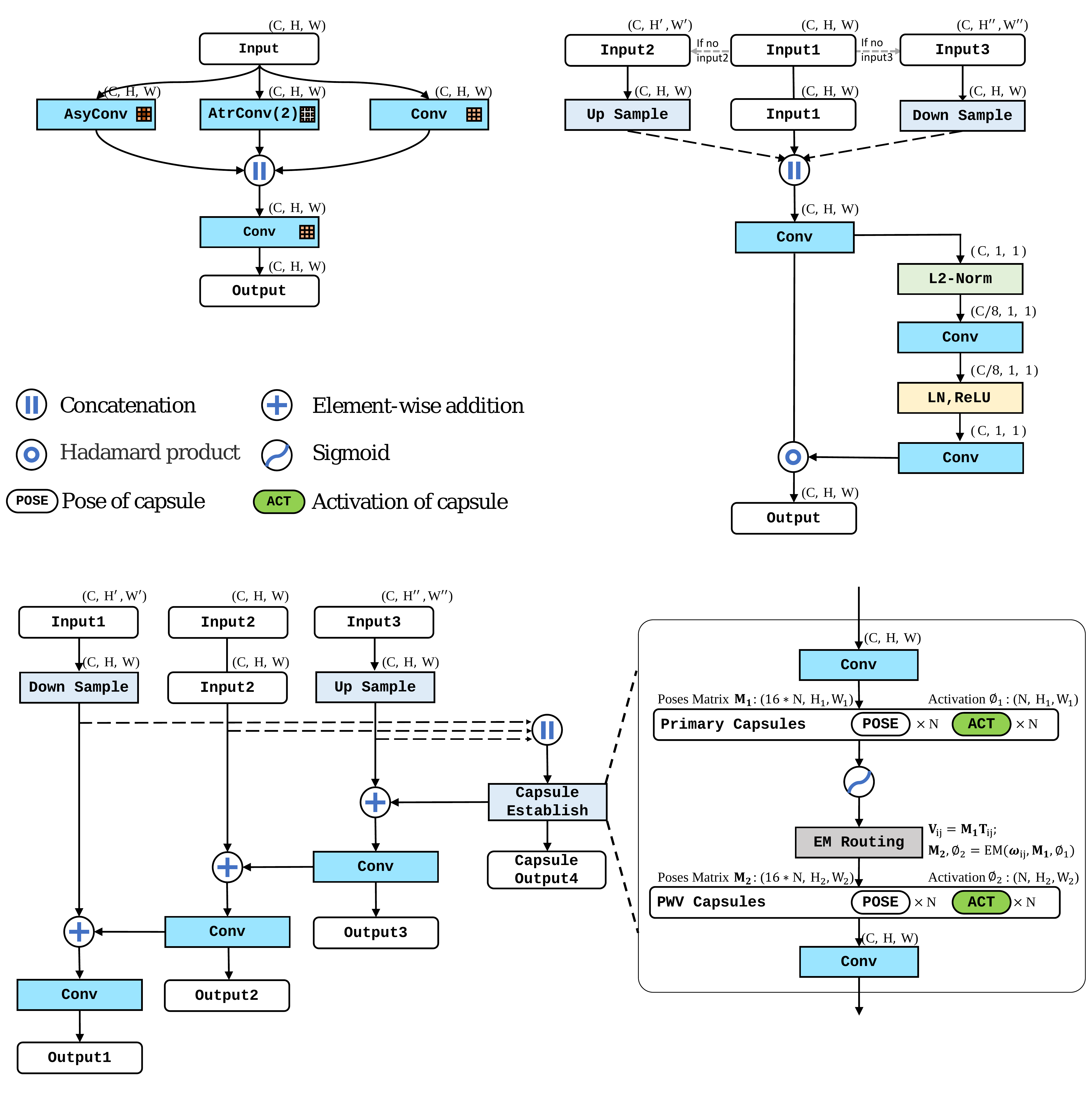} 
     \put(19.8,68.8){\footnotesize{(a) DFA}}
     \put(69.00,47.8){\footnotesize{(b) ICE}}
     \put(45.5,2.5){\footnotesize{(c) PWV}}
	\end{overpic}
	\vspace{-10pt}
	\caption{Details of the proposed modules. (a) The diverse feature aggregation (DFA) component combines different convolutional kernels to enhance the representational ability. (b) The integrity channel enhancement \Rev{(ICE)} component, mines integrity information along the channel dimension. (c) The part-whole verification \Rev{(PWV)} component is designed for modeling the relation between the parts and the whole object. ``AsyConv'', ``AtrConv'', and ``Conv'' means \emph{asymmetric block}~\cite{ding2019acnet}, \emph{atrous block}~\cite{chen2017deeplab}, and conv block, respectively. All these blocks include convolution, BatchNorm and ReLU components. ``EM Routing'' means the expectation-maximum routing mechanism~\cite{hinton2018matrix}. C, H, W denote the channel number, height, and width of the feature tensor, respectively. 
	}
	\label{fig:modules}
\end{figure*}

\Rev{Recent works~\cite{chen2017rethinking, szegedy2016rethinking, zhao2017pyramid} have demonstrated that enriching the receptive fields of the convolution kernel can help the network learn features that capture different object sizes}. In this work, we go one step further and incorporate convolution kernels with different shapes to deal with the shape diversity of different objects. Specifically, as shown in \figref{fig:modules}-(A), we introduce the novel DFA module to enhance the diversity of the extracted multi-level features by using three kinds of convolutional blocks with different kernel sizes and shapes. Technically, we utilize a practical combination of the asymmetric convolution~\cite{ding2019acnet}, atrous convolution~\cite{chen2017deeplab}, and original convolution operations to capture diverse spatial features. The overall procedure is summarized as follows:
\Rev{
 \begin{equation}\begin{array}{l}
 \mathbf{F}^{(i)}_{dfa}=\operatorname{Concat}\left[\mathcal{X}_{asy}(\mathbf{F}^{(i)}_{bkb}), \mathcal{X}_{atr}^{2}(\mathbf{F}^{(i)}_{bkb}), \mathcal{X}_{ori}(\mathbf{F}^{(i)}_{bkb})\right],
 \end{array}\end{equation}}

\noindent where $\mathbf{F}^{(i)}_{dfa}$ denotes the features produced by the above process, $\mathcal{X}_{*}$ denotes different types of blocks (\ie, asymmetric, atrous, original convolutions), and \Rev{$\operatorname{Concat}[\cdot]$} is the concatenation operation. 

Note that we use $\mathcal{X}_{atr}^{r}$ to denote atrous convolution operations with different dilation rates $r$, for example, $\mathcal{X}_{atr}^{2}$ is the atrous convolution operation with the dilation rate of 2, and use the asymmetric convolution ($\mathcal{X}_{asy}$) operation with a crux-shape \cite{ding2019acnet}. Specially, $\mathcal{X}_{asy}$ contains three layers, one with a normal $3 \times 3$ square kernel \Rev{$\textbf{K}_{3\times3}$}, one with a horizontal $1 \times 3$ kernel \Rev{$\textbf{K}_{1\times3}$}, and one with a vertical $3 \times 1$ kernel \Rev{$\textbf{K}_{3\times1}$}, and all of them are shared in the same sliding window. It can be described as:
\begin{equation}\begin{array}{l}
\mathcal{X}_{asy}(\mathbf{I})=\left(\mathbf{I} \star \mathbf{K}_{3\times3}\right) \oplus \left(\mathbf{I} \star \mathbf{K}_{1\times3}\right) \oplus \left(\mathbf{I} \star \mathbf{K}_{3\times1}\right),
\end{array}\end{equation}
where $\star$ is the 2D convolutional operator, $\oplus$  is the element-wise addition, and $\textbf{I}$ denotes the input feature.

In such a way, our DFA module can enrich the feature space by fusing the learned knowledge from the crux kernel, dilated kernel, and normal kernel in the first stage. 
As a result, DFA can cover different salient regions in various contexts, by enhancing integrity. We mark the features processed by DFA  as $\mathcal{F}_{dfa}=\{\mathbf{F}^{(\Red{1})}_{dfa}, \mathbf{F}^{(\Red{2})}_{dfa}, \mathbf{F}^{(\Red{3})}_{dfa}, \mathbf{F}^{(\Red{4})}_{dfa}\}$.

\begin{figure*}[t!]
	\centering
	\begin{overpic}[width=1.0\textwidth]{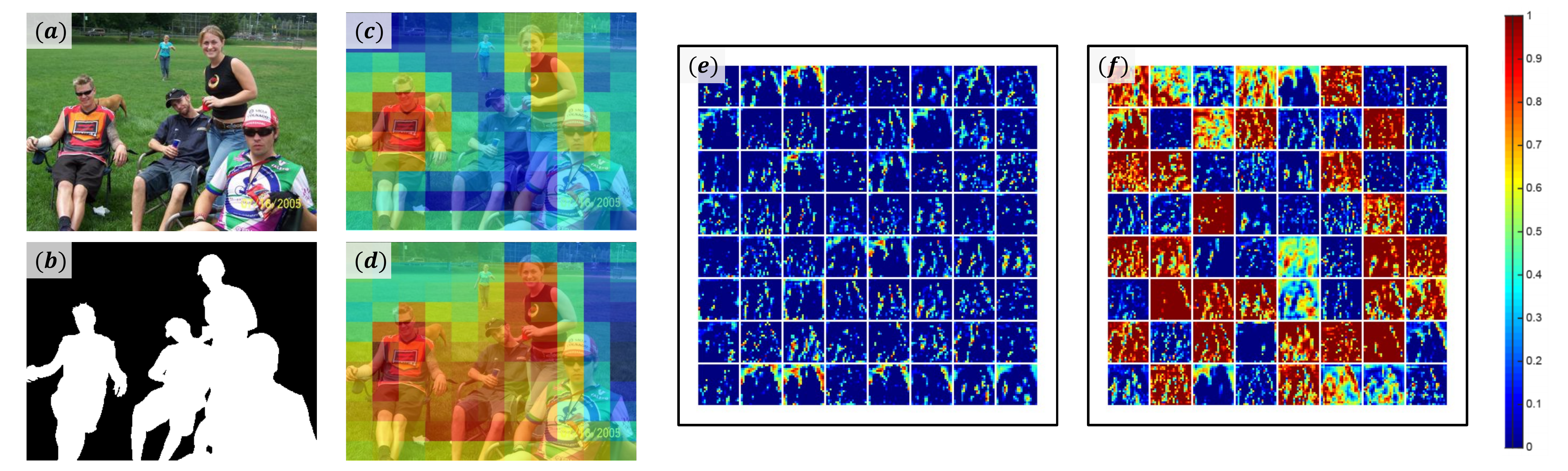} 
	\end{overpic}
	\vspace*{-20pt}
	\caption{Visual comparison of heatmaps before and after using our ICE. %
	(a) Input image. (b) Ground truth. Feature heatmaps (c) before and (d) after our ICE module. 
	Channel visualizations (e) before and (f) after the ICE module. 
	Our ICE module helps the network focus more on the integral salient regions 
	and distinguish the foreground regions from the background. 
	The feature heatmap presented in \Red{(d)} clearly demonstrates the ability of our ICE module to capture the integrity representation at both macro- and micro levels. 
	The channel visualizations are generated by squeezing all channels and then we use Matplotlib's pseudo-color `jet' for colorization. Zoom in for better viewing.
	}\label{fig:ice_visual}
\end{figure*}


\subsection{Integrity Channel Enhancement}\label{ICE} 

Several recent studies~\cite{hu2020squeeze,woo2018cbam,cao2019gcnet,yang2020gated} have achieved promising visual categorization results by using the spatial or channel attention mechanism. Though these methods are driven by various motivations, they all essentially aim to build the correspondence between different features to highlight the most significant object parts. 
However, how to mine the integrity information hidden in different channel of features remains under studied.
To address this issue, we propose a simple ICE module to further mine the relations within different channels, and enhance the channels that highlight the potential integral targets.

We consider multi-scale information from every three adjacent features. First, we re-scale the next and previous feature levels and use upsampling and downsampling operations to adjust them to the spatial resolution of $H \times W$. 
\Rev{Then, we generate the fusion maps $\mathbf{F}^{(i)}_{fuse}$ by concatenating the three input features:}
\Rev{\begin{equation}\begin{array}{l}
\mathbf{F}^{(i)}_{fuse}=\operatorname{Concat}\left[\mathbf{F}^{(i-1)}_{dfa}, \mathbf{F}^{(i)}_{dfa}, \mathbf{F}^{(i+1)}_{dfa}\right].
\end{array}
\end{equation}}


\noindent 

After that, we extract the integrity embedding $\mathbf{I}_{e m b}^{(i)}$ by applying the $l_2$ norm on $\mathbf{F}_{fuse}^{(i)}$. 
Next, to further integrate the integrity information, we use a parameter-efficient bottleneck design to learn $\mathbf{I}_{emb}^{(i)}$. 
\Red{As the channel transform would slightly increase the difficulty of optimization, we add layer normalization inside two convolution layers (before ReLU) to facilitate optimization, similar to the design used in~\cite{cao2019gcnet}:}
\Rev{
\begin{equation}
\mathbf{F}_{ice}^{(i)} = \mathbf{F}_{fuse}^{(i)} \otimes \mathcal{X}_{ori} (\operatorname{ReLU} ( \operatorname{LN} ( \mathcal{X}_{ori}(\mathbf{I}_{emb}^{(i)})))),
\end{equation}
}
\vspace{-5pt}
\\
\Rev{where $\otimes$ is the element-wise multiplication operation and $\operatorname{LN}$ means layer normalization.}

By using the proposed ICE module, the channels with better integrity can be effectively enhanced. As can be seen in \figref{fig:ice_visual}, \Rev{after feeding the features into our ICE, the foreground region is noticeably distinguished from the background,} and the features produced by ICE tend to highlight the integral objects at both the micro and macro levels. In our implementation, if there are not enough multi-level features input in the first and last levels, we will fill the input with the features from the current level. \Rev{Besides, we use two ICE modules with shared parameters to help our ICON model integrate cues at multiple levels.}

\subsection{Part-Whole Verification}\label{PWV}

The PWV module aims to enhance the learned integrity features by measuring the agreement between object parts and the whole salient region. To achieve this goal, we adopt a capsule network~\cite{sabour2017dynamic,hinton2018matrix}, which has been proved to be effective in modeling part-whole relationships. Motivated by the success of the prior work SegCaps~\cite{lalonde2018capsules}, we embed the capsule network into our \ourmodel. 
\Red{In PWV, one key issue is how to assign votes from the low-level to the high-level capsules. As the high-level capsules need to form the whole object representation by aggregating the object parts from the relevant low-level capsules, we use EM routing~\cite{hinton2018matrix} to model the association between the low-level and high-level capsules in a clustering-like manner. }
The inputs of PWV are \Rev{three} different ICE features ($\mathcal{F}_{ice}$).  Specifically, we first reduce the ICE features at each level to a united resolution, \ie, ~$22 \times 22$, in order to reduce the computational costs.

Next, we build our {primary capsules}. To be specific, we use eight pose vectors to build a pose matrix \textbf{M}, and an activation $ \phi \in \left[0,1\right]$ to represent each capsule. The pose matrix contains the instantiated parameters to reflect the properties of object parts or the whole object, while the activation represents the existence probability of the object. The capsules from the primary capsule layer pass information to those in the next PWV capsule layer through a routing-by-agreement mechanism. Specifically, when the capsules from a lower layer produce votes for the capsules in a higher level, the votes $\omega_{i j}$ are obtained by a matrix multiplication operation between the learned transformation matrices $\textbf{T}_{i j}$ and the lower-level pose matrix $\textbf{M}_i$, where $i$ and $j$ are the indices of the lower- and higher level capsules, respectively. Once these votes are obtained, they are used in the EM routing algorithm~\cite{hinton2018matrix} to produce the higher-level capsule $\textbf{C}_j$ with the pose matrices $\textbf{M}_{j}$ and the activation $\phi_{j}$. After that, we obtain the part-whole verified features. 
Subsequently, element-wise addition and upsampling operations are introduced to fuse these part-whole verified features at adjacent levels in a bottom-up manner, which encourages cooperation among multi-scale features. After the PWV module, the model outputs $\mathcal{F}_{pwv}=\{\mathbf{F}^{({1})}_{pwv}, \mathbf{F}^{({2})}_{pwv}, \mathbf{F}^{({3})}_{pwv}, \mathbf{F}^{({4})}_{pwv\_cap}\}$.

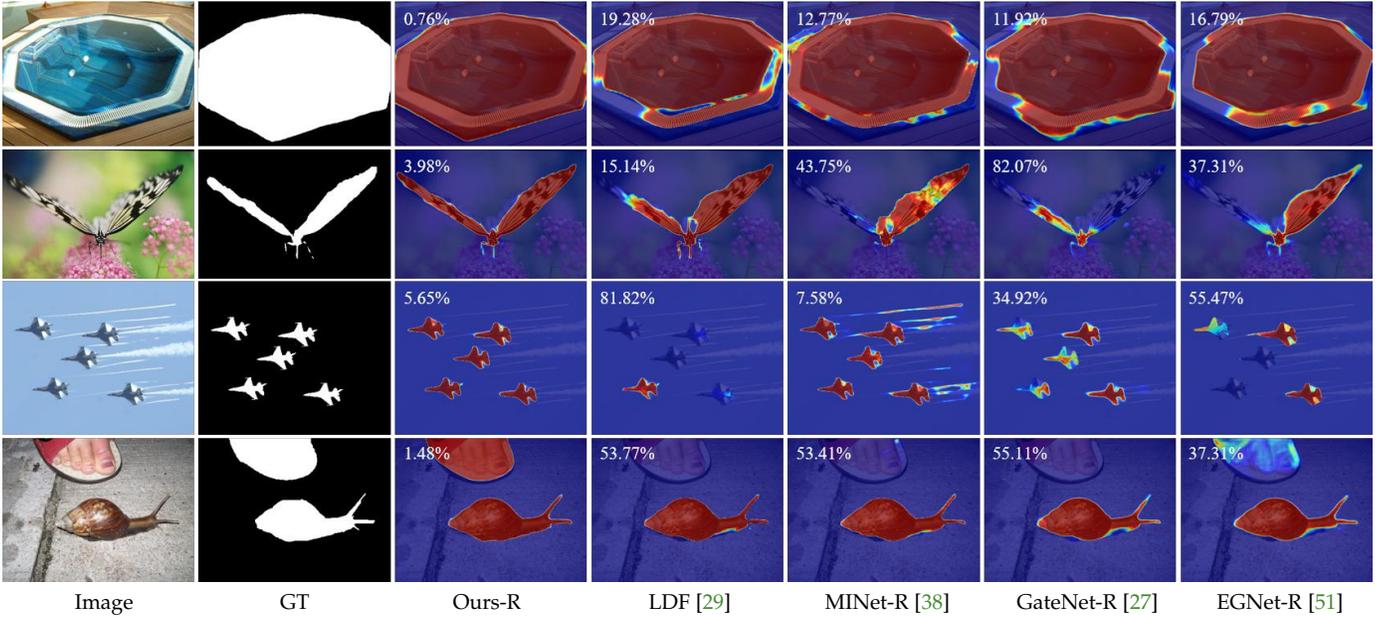
\begin{figure*}[t!]
	\centering
	\begin{overpic}[width=\textwidth]{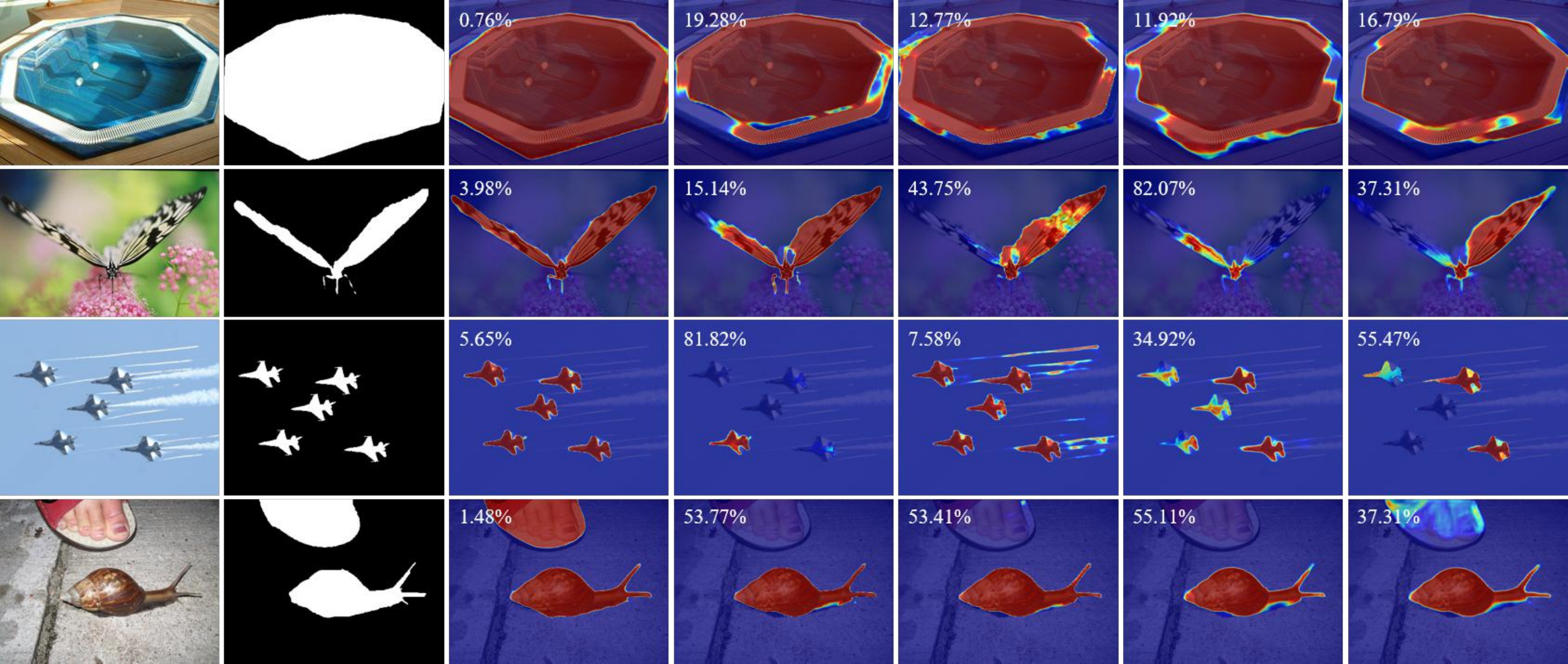} 
	\put(5.3,-2){\footnotesize{Image}}
	\put(20.3,-2){\footnotesize{GT}}
	\put(32.9,-2){\footnotesize{Ours-R}}
	\put(47.2,-2){\footnotesize{LDF~\cite{wei2020label}}}
    \put(60.1,-2){\footnotesize{MINet-R~\cite{pang2020multi}}}
    \put(74.1,-2){\footnotesize{GateNet-R~\cite{zhao2020suppress}}}
    \put(88.7,-2){\footnotesize{EGNet-R~\cite{zhao2019egnet}}}
	\end{overpic}
	\caption{\Rev{Visual comparison at different false negative ratios (FNRs). Note that the FNRs heavily reflect the \textit{integrity} scores.}}
\label{fig:fnr2}
\end{figure*}

\subsection{Supervision Strategy}\label{FFS}
In this work, in addition to the BCE loss, we also use the IoU loss~\cite{qin2019basnet, mattyus2017deeproadmapper}. Specifically, the overall loss of the proposed \ourmodel~is formulated as $
\mathcal{L}_{CPR}\left(\textit{P}, \textit{G}\right) $, 
where $\textit{P}$ is the generated saliency prediction map, and $\textit{G}$ is the ground truth saliency map. $\mathcal{L}_{CPR} $ incorporates the cooperative BCE loss and IoU loss, \ie,~$\mathcal{L}_{CPR}=\mathcal{L}_{BCE}+\mathcal{L}_{IoU}$. Specifically, $\mathcal{L}_{BCE}$ is formulated as follows:   

\begin{footnotesize}     
	\begin{equation}\small
	\begin{aligned}
	\mathcal{L}_{BCE}= &-\sum_{\textit{x}=1}^{\textit{H}} \sum_{\textit{y}=1}^{\textit{W}}[\textit{G}(\textit{x}, \textit{y}) \log (\textit{P}(\textit{x}, \textit{y})) \\ &+(1-\textit{G}(\textit{x}, \textit{y})) \log (1-\textit{P}(\textit{x}, \textit{y}))],
	\end{aligned}
	\label{equ:bce}
	\end{equation}
\end{footnotesize}
\vspace{-5pt}
\\
\noindent where $\textit{W}$ and $\textit{H}$ are the width and height of the images, respectively. Meanwhile, $\textit{L}_{IoU}$ is defined as: 

\begin{footnotesize} 
	\begin{equation}\small
	\mathcal{L}_{IoU}=1-\frac{\sum_{\textit{x}=1}^{\textit{H}} \sum_{\textit{y}=1}^{\textit{W}} \textit{P}(\textit{x}, \textit{y}) \textit{G}(\textit{x}, \textit{y})}{\sum_{\textit{x}=1}^{\textit{H}} \sum_{\textit{y}=1}^{\textit{W}}[\textit{P}(\textit{x}, \textit{y})+\textit{G}(\textit{x}, \textit{y})-\textit{P}(\textit{x}, \textit{y}) \textit{G}(\textit{x}, \textit{y})]},
	\label{equ:iou}
	\end{equation}
\end{footnotesize}
\vspace{-5pt}
\\
\noindent where $\textit{G}(\textit{x}, \textit{y})$ \Rev{and $\textit{P}(\textit{x}, \textit{y})$ are the ground truth label and the predicted saliency label at the location $(\textit{x}, \textit{y})$, respectively.}
\Red{During training, we use the multi-level supervision strategy widely used in this field~\cite{liu2019simple, GCPANet, F3Net, pang2020multi}. 
Apart from using four features from $\mathcal{F}_{pwv}$, we fuse $\mathbf{F}^{({1})}_{pwv}$ and $\mathbf{F}^{({1})}_{ice}$ by dot-product, which is used as an extra feature for supervision. This feature is also used to generate the final prediction results during the inference period.  To match the ground-truth maps in both training and inference stages, the features' channel will be reduced to 1-dimension, and the spatial size will be recovered as the same as the input. 
}


\section{Experiments}\label{sec4}
\subsection{Datasets}
We train our \ourmodel~on the \textbf{DUTS-TR}~\cite{wang2017learning}, which is commonly used for the SOD task and contains 10,553 images. Then, we evaluate all the models on seven popular datasets: \textbf{ECSSD}~\cite{yan2013hierarchical}, \textbf{HKU-IS}~\cite{li2015visual}, \textbf{OMRON}~\cite{yang2013saliency}, \textbf{PASCAL-S} \cite{li2014secrets}, \textbf{DUTS-TE}~\cite{wang2017learning},  \textbf{SOD}~\cite{movahedi2010design} and attribute-based \textbf{SOC}~\cite{fan2018SOC}, which are all annotated with pixel-level labels. Specifically, ECSSD 
is made up of 1,000 images with meaningful semantics. HKU-IS includes 4,447 images, containing multiple foreground objects. OMRON consists of 5,168 images with at least one object. These objects are usually structurally complex. PASCAL-S was built from a dataset originally used for semantic segmentation, and it consists of 850 challenging images. DUTS is a relatively large dataset with two subsets. The 10,553 images in DUTS-TR are used for training, and the 5,019 images in DUTS-TE are employed for testing. SOD 
includes 300 very challenging images. SOC contains images from complicated scenes, which are more challenging than those in the other six datasets.

\begin{table*}[t!]
	\centering
	\footnotesize
	\renewcommand{\arraystretch}{0.9}
	\renewcommand{\tabcolsep}{0.0001pt}
	\caption{Quantitative results on six datasets. 
	The best results are shown in \textbf{bold}.
	The symbols ``$\uparrow$''/``$\downarrow$'' mean that a higher/lower score is better. '-V/VGG-Based': VGG16~\Rev{\cite{simonyan2014very}}, '-R/ResNet-Based': ResNet50~\Rev{\cite{he2016deep}}, '-P': PVTv2-1K~\Rev{\cite{wang2021pvtv2}}, '-S': Swin-B-22k~\Rev{\cite{liu2021swin}}, '-M': CycleMLP-B4~\Rev{\cite{chen2021cyclemlp}}. }
   \begin{tabular}{ccccccccccccccccccccccccccc}
\hline
\multicolumn{3}{c|}{\textbf{Summary}} & \multicolumn{4}{c|}{\textbf{ECSSD}~\cite{yan2013hierarchical}} & \multicolumn{4}{c|}{\textbf{PASCAL-S}~\cite{li2014secrets}} & \multicolumn{4}{c|}{\textbf{DUTS}~\cite{wang2017learning}} & \multicolumn{4}{c|}{\textbf{HKU-IS}~\cite{li2015visual}} & \multicolumn{4}{c|}{\textbf{OMRON}~\cite{yang2013saliency}} & \multicolumn{4}{c}{\textbf{SOD}~\cite{movahedi2010design}} \\ \hline
\multicolumn{1}{l|}{Method} & \multicolumn{1}{c|}{MACs} & \multicolumn{1}{c|}{Params} & $S_{m} \uparrow$ & $E_\xi^{m}\uparrow$ & $F_{\beta}^{w}\uparrow$ & \multicolumn{1}{c|}{\emph{M}$\downarrow$}  & $S_{m} \uparrow$ & $E_\xi^{m}\uparrow$ & $F_{\beta}^{w}\uparrow$ & \multicolumn{1}{c|}{\emph{M}$\downarrow$}  & $S_{m} \uparrow$ & $E_\xi^{m}\uparrow$ & $F_{\beta}^{w}\uparrow$ & \multicolumn{1}{c|}{\emph{M}$\downarrow$}  & $S_{m} \uparrow$ & $E_\xi^{m}\uparrow$ & $F_{\beta}^{w}\uparrow$ & \multicolumn{1}{c|}{\emph{M}$\downarrow$} & $S_{m} \uparrow$ & $E_\xi^{m}\uparrow$ & $F_{\beta}^{w}\uparrow$ & \multicolumn{1}{c|}{\emph{M}$\downarrow$}  & $S_{m} \uparrow$ & $E_\xi^{m}\uparrow$ & $F_{\beta}^{w}\uparrow$ & \multicolumn{1}{c}{\emph{M}$\downarrow$}\\ \hline
\multicolumn{27}{c}{VGG16-Based Methods} \\ \hline
\multicolumn{1}{l|}{RAS} & \multicolumn{1}{c|}{-} & \multicolumn{1}{c|}{-} & .893 & .914 & .857 & \multicolumn{1}{c|}{.056} & .799 & .835 & .731 & \multicolumn{1}{c|}{.101} & .839 & .871 & .740 & \multicolumn{1}{c|}{.059} & .887 &.920 & .843 & \multicolumn{1}{c|}{.045} & .814 & .843 & .695 & \multicolumn{1}{c|}{.062} & .767 & .791 & .718 &.123\\
\multicolumn{1}{l|}{CPD} & \multicolumn{1}{c|}{59.46} & \multicolumn{1}{c|}{29.23} &.910 & .938 & .895 & \multicolumn{1}{c|}{.040} & .845 & .882 & .796 & \multicolumn{1}{c|}{.072} & .867 & .902 & .800 & \multicolumn{1}{c|}{.043} & .904 & .940 & .879 & \multicolumn{1}{c|}{.033} & .818 & .845 & .715 & \multicolumn{1}{c|}{.057} & .771 & .787 & .718 & 113 \\
\multicolumn{1}{l|}{EGNet} & \multicolumn{1}{c|}{149.89} & \multicolumn{1}{c|}{108.07} & \textbf{.919}  & .936 & .892 & \multicolumn{1}{c|}{.041} & .848 & .877 & .788 & \multicolumn{1}{c|}{.077} & {.878} & .898 & .797 & \multicolumn{1}{c|}{.044} & .910 & .938 & .875 & \multicolumn{1}{c|}{.035} & \textbf{.836} & {.853} & .728 & \multicolumn{1}{c|}{\textbf{.057}} & .788 & .803 &.736 & .110 \\ 
\multicolumn{1}{l|}{ITSD} & \multicolumn{1}{c|}{14.61} & \multicolumn{1}{c|}{17.08} & .914 & .937 & .897 & \multicolumn{1}{c|}{.040} & .856 & .891 &{.811}  & \multicolumn{1}{c|}{{.068}} & .877 & .905 &{.814}  & \multicolumn{1}{c|}{.042} & .906 & .938 & .881 & \multicolumn{1}{c|}{.035} & {.829} & {.853} & {.734} & \multicolumn{1}{c|}{.063} & {.797} & {.826} & {.764} & {.098} \\ 
\multicolumn{1}{l|}{MINet} & \multicolumn{1}{c|}{94.11} & \multicolumn{1}{c|}{47.56} &\textbf{.919}  &{.943}  &\textbf{.905}  & \multicolumn{1}{c|}{\textbf{.036}} & .854 & {.893} & .808 & \multicolumn{1}{c|}{\textbf{.064}} & .875 & {.907} & .813 & \multicolumn{1}{c|}{\textbf{.039}} & {.912} & {.944} & {.889} & \multicolumn{1}{c|}{\textbf{.031}} & .822 & .846 & .718 & \multicolumn{1}{c|}{\textbf{.057}} & - & - & - & - \\ 
\multicolumn{1}{l|}{GateNet} & \multicolumn{1}{c|}{108.34} & \multicolumn{1}{c|}{100.02} & .917 & .932 & .886 & \multicolumn{1}{c|}{.041} & {.857}  & .886 & .797 & \multicolumn{1}{c|}{{.068} } & .870 & .893 & .786 & \multicolumn{1}{c|}{.045} & .910 & .934 & .872 & \multicolumn{1}{c|}{.036} & .821 & .840 &.703 & \multicolumn{1}{c|}{.061} & - & - & - & - \\ 
\rowcolor{mygray}
\multicolumn{1}{l|}{Ours-V} & \multicolumn{1}{c|}{64.90} & \multicolumn{1}{c|}{19.17} & \textbf{.919} & \textbf{.946} & \textbf{.905} & \multicolumn{1}{c|}{\textbf{.036}} & \textbf{.861} & \textbf{.902} & \textbf{.820} & \multicolumn{1}{c|}{\textbf{.064}} & \textbf{.878} & \textbf{.915} & \textbf{.822} & \multicolumn{1}{c|}{{.043}} & \textbf{.915} & \textbf{.950} & \textbf{.895} & \multicolumn{1}{c|}{{.032}} & .833 & \textbf{.865} &\textbf{.743} & \multicolumn{1}{c|}{.065} & \textbf{.814} & \textbf{.848} & \textbf{.784} & \textbf{.089} \\
\hline
\multicolumn{27}{c}{ResNet50-Based Methods} \\ \hline
\multicolumn{1}{l|}{CondInst} & \multicolumn{1}{c|}{-} & \multicolumn{1}{c|}{-} & .721 & .717 & .603 & \multicolumn{1}{c|}{.115} & .813 & .852 &.757 & \multicolumn{1}{c|}{.084} & .760 & .764& .631 & \multicolumn{1}{c|}{.070} & .748 & .754 & .648 & \multicolumn{1}{c|}{.093} & .646 & .629 & .433 & \multicolumn{1}{c|}{.114} & .670 & .657 & .535 & .148 \\ 
\multicolumn{1}{l|}{PointRend} & \multicolumn{1}{c|}{-} & \multicolumn{1}{c|}{-} & .753 & .766 & .667 & \multicolumn{1}{c|}{.111} & .810 & .850 &.763 & \multicolumn{1}{c|}{.099} & .774 & .794 & .667 & \multicolumn{1}{c|}{.081} & .784 & .805 & .715 & \multicolumn{1}{c|}{.091} & .651 & .647 & .453 & \multicolumn{1}{c|}{.125} & .693 & .793 & .589 & .141 \\ 
\multicolumn{1}{l|}{PiCANet} & \multicolumn{1}{c|}{54.05} & \multicolumn{1}{c|}{47.22} & .917 & .925 & .867 & \multicolumn{1}{c|}{.046} & .854 & .870 & .772 & \multicolumn{1}{c|}{.076} & .869 & .878 & .754 & \multicolumn{1}{c|}{.051} & .904 & .916 & .840 & \multicolumn{1}{c|}{.043} & .832 & .836 & .695 & \multicolumn{1}{c|}{.065} & .793 & .799 & .722 & .103 \\
\multicolumn{1}{l|}{AFNet} & \multicolumn{1}{c|}{21.66} & \multicolumn{1}{c|}{35.95} & .913 & .935 & .886 & \multicolumn{1}{c|}{.042} & .849 & .883 & .797 & \multicolumn{1}{c|}{.070} & .867 & .893 & .785 & \multicolumn{1}{c|}{.046} & .905 & .935 & .869 & \multicolumn{1}{c|}{.036} & .826 & .846 & .717 & \multicolumn{1}{c|}{.057} & - & - & - & - \\
\multicolumn{1}{l|}{BASNet} & \multicolumn{1}{c|}{127.36} & \multicolumn{1}{c|}{87.06} & .916 & .943 & .904 & \multicolumn{1}{c|}{.037} & .838 & .879 & .793 & \multicolumn{1}{c|}{.076} & .866 & .895 & .803 & \multicolumn{1}{c|}{.040} & .909 & .943 & .889 & \multicolumn{1}{c|}{.032} & .836 & .865 & .751 & \multicolumn{1}{c|}{.056} & .772 & .801 & .728 & .112 \\
\multicolumn{1}{l|}{CPD} & \multicolumn{1}{c|}{17.77} & \multicolumn{1}{c|}{47.85} & .918 & .942 &.898 & \multicolumn{1}{c|}{.037} & .848 & .882 & .794 & \multicolumn{1}{c|}{.071} & .869 & .898 & .795 & \multicolumn{1}{c|}{.043} & .905 & .938 & .875 & \multicolumn{1}{c|}{.034} & .825 & .847 & .719 & \multicolumn{1}{c|}{.056} & .771 &.782 & .713 & .110\\
\multicolumn{1}{l|}{EGNet} & \multicolumn{1}{c|}{157.21} & \multicolumn{1}{c|}{111.69} & .925 & .943 & .903 & \multicolumn{1}{c|}{.037} & .852 & .881 & .795 & \multicolumn{1}{c|}{.074} & {.887} & .907 & .815 & \multicolumn{1}{c|}{.039} & .918 & .944 & .887 & \multicolumn{1}{c|}{.031} & {.841} & .857 & .738 & \multicolumn{1}{c|}{\textbf{.053}} & .807 & .822 & .767 & .097 \\
\multicolumn{1}{l|}{SCRN} & \multicolumn{1}{c|}{15.09} & \multicolumn{1}{c|}{25.23} & .927 & .939 & .900 & \multicolumn{1}{c|}{.037} & \textbf{.869} & .892 & .807 & \multicolumn{1}{c|}{{.063}} & .885 & .900 & .803 & \multicolumn{1}{c|}{.040} & .916 & .935 & .876 & \multicolumn{1}{c|}{.034}  & .837 & .848 & .720 & \multicolumn{1}{c|}{.056} & - & - & - & - \\
\multicolumn{1}{l|}{F3Net} & \multicolumn{1}{c|}{16.43} & \multicolumn{1}{c|}{25.54} & .924 & .948 & {.912} & \multicolumn{1}{c|}{{.033}} & .861 &  {.898} & {.816} & \multicolumn{1}{c|}{\textbf{.061}} & \textbf{.888} & {.920} & {.835} & \multicolumn{1}{c|}{\textbf{.035}} & .917 & {.952}  & {.900} & \multicolumn{1}{c|}{\textbf{.028}} & .838 & .864 & .747 & \multicolumn{1}{c|}{\textbf{.053}} & .806 & .834 & .775 & {.091} \\ 
\multicolumn{1}{l|}{ITSD} & \multicolumn{1}{c|}{15.96} & \multicolumn{1}{c|}{26.47} & .925 & .947 & .910 & \multicolumn{1}{c|}{.034} & .859 & .894 & .812 & \multicolumn{1}{c|}{.066} & .885 & .913 & .823 & \multicolumn{1}{c|}{.041} & .917 & .947 & .894 & \multicolumn{1}{c|}{.031} & .840 & {.865} & {.750} & \multicolumn{1}{c|}{.061} &{.809}  &{.836}  &{.777}  & .093 \\
\multicolumn{1}{l|}{MINET} & \multicolumn{1}{c|}{87.11} & \multicolumn{1}{c|}{126.38} & .925 & {.950} & .911 & \multicolumn{1}{c|}{{.033}} & .856 & .896 & .809 & \multicolumn{1}{c|}{.064} & .884 & .917 & .825 & \multicolumn{1}{c|}{{.037}} & {.919} & {.952} & .897 & \multicolumn{1}{c|}{{.029}} & .833 & .860 & .738 & \multicolumn{1}{c|}{.056} & - & - & - & - \\
\multicolumn{1}{l|}{GateNet} & \multicolumn{1}{c|}{162.13} & \multicolumn{1}{c|}{128.63} & .920 & .936 & .894 & \multicolumn{1}{c|}{.040} & .858 & .886 & .797 & \multicolumn{1}{c|}{.067} & .885 & .906 & .809 & \multicolumn{1}{c|}{.040} & .915 & .937 & .880 & \multicolumn{1}{c|}{.033} & .838 & .855 & .729 & \multicolumn{1}{c|}{.055} & - & - & - & - \\ 
\rowcolor{mygray}
\multicolumn{1}{l|}{Ours-R} & \multicolumn{1}{c|}{20.91} & \multicolumn{1}{c|}{33.09} & \textbf{.929} & \textbf{.954} & \textbf{.918} & \multicolumn{1}{c|}{\textbf{.032}} & {.861} & \textbf{.899} & \textbf{.818} & \multicolumn{1}{c|}{{.064}} & \textbf{.888} & \textbf{.924} & \textbf{.836} & \multicolumn{1}{c|}{.037} & \textbf{.920} & \textbf{.953} & \textbf{.902} & \multicolumn{1}{c|}{{.029}} & \textbf{.844} &\textbf{.876}& \textbf{.761} & \multicolumn{1}{c|}{.057} & \textbf{.824} & \textbf{.854} & \textbf{.794} & \textbf{.084} \\ 
\hline
\multicolumn{27}{c}{Transformer-Based Methods} \\ \hline
\multicolumn{1}{l|}{VST} & \multicolumn{1}{c|}{23.16} & \multicolumn{1}{c|}{44.63} & .932 & .951 & .910 & \multicolumn{1}{c|}{.033} & .872 & .902 & .816 & \multicolumn{1}{c|}{.061} & .896 & .919 & .828 & \multicolumn{1}{c|}{.037} & .928 & .952 & .897 & \multicolumn{1}{c|}{.029} & .850 & .871 & .755 & \multicolumn{1}{c|}{.058} & .820 & .846 & .778 & .086 \\
\rowcolor{mygray}
\multicolumn{1}{l|}{Ours-P} & \multicolumn{1}{c|}{34.70} & \multicolumn{1}{c|}{65.68} & {.940} & {.964} & {.933} & \multicolumn{1}{c|}{.024} & {.882} & {.921} &{.847} & \multicolumn{1}{c|}{{.051}} & \textbf{.917} & {.950} & {.882} & \multicolumn{1}{c|}{\textbf{.022}} & \textbf{.935} & {.967} & \textbf{.925} & \multicolumn{1}{c|}{\textbf{.022}} & .865 & {.896} & {.793} & \multicolumn{1}{c|}{{.047}} & {\textbf{.832}} & \textbf{.864} & \textbf{.813} & \textbf{.078} \\ 
\rowcolor{mygray}
\multicolumn{1}{l|}{Ours-S} & \multicolumn{1}{c|}{52.59} & \multicolumn{1}{c|}{94.30} & \textbf{.941} & \textbf{.966} & \textbf{.936} & \multicolumn{1}{c|}{\textbf{.023}} & \textbf{.885} & \textbf{.924} &\textbf{.854} & \multicolumn{1}{c|}{\textbf{.048}} & {\textbf{.917}} & \textbf{.954} & \textbf{.886} & \multicolumn{1}{c|}{.025} & {\textbf{.935}} & \textbf{.968} & \textbf{.925} & \multicolumn{1}{c|}{\textbf{.022}} & \textbf{.869} & \textbf{.900} & \textbf{.804} & \multicolumn{1}{c|}{\textbf{.043}} & .825 & .856 & .802 & .083 \\ 
\hline
\multicolumn{27}{c}{MLP-Based Methods} \\ \hline
\rowcolor{mygray}
\multicolumn{1}{l|}{Ours-M} & \multicolumn{1}{c|}{26.13} & \multicolumn{1}{c|}{54.92} & .940 & .964 & .934 & \multicolumn{1}{c|}{.025} & .873 & .912 & .838 & \multicolumn{1}{c|}{.056} & .909 & .942 & .874 & \multicolumn{1}{c|}{.029} & .935 & .966 & .926 & \multicolumn{1}{c|}{.022} & .855 & .886 & .783 & \multicolumn{1}{c|}{.051} & .821 & .853 & .803 & .081 \\
\hline
\end{tabular}
\label{tab:comp1}
\end{table*}

\begin{figure*}[t!]
	\centering
	\begin{overpic}[width=0.95\textwidth]{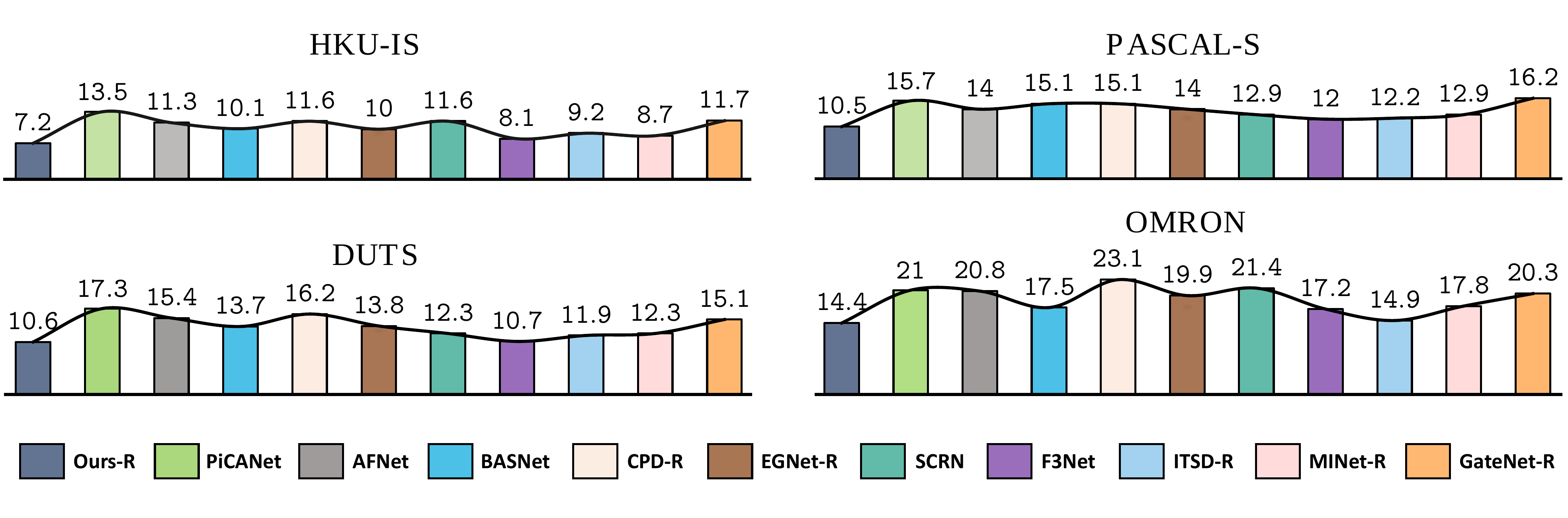} 
	\end{overpic}
	\vspace*{-15pt}
	\caption{\Rev{FNR statistics of 11 methods on four different datasets.}}
\label{fig:fnr1}
\end{figure*}

\subsection{Implementation Details}\label{imp_details}
We run all experiments on the publicly available Pytorch 1.5.0 platform. An eight-core PC with an Intel Core i7-9700K CPU (with 4.9GHz Turbo boost), 16GB 3000 MHz RAM and an RTX 2080Ti GPU card (with 11GB memory) is used for both training and testing. During network training, each image is first resized to \Rev{352$\times$352~(for the VGG~\cite{simonyan2014very}/ResNet~\cite{he2016deep}/PVT~\cite{wang2021pyramid,wang2021pvtv2} backbones) or 384$\times$384~(for Swin~\cite{liu2021swin}/CycleMLP~\cite{chen2021cyclemlp}),} and data augmentation methods such as normalizing, cropping and flipping, are used. Some encoder parameters are initialized from \Rev{VGG-16, ResNet-50}, \Rev{PVTv2, Swin-B and CycleMLP-B4}. We initialize some layers of PWV by zeros or ones, while other convolutional layers are initialized based on \cite{he2015delving}. We use the SGD optimizer~\cite{bottou2012stochastic} to train our network, and set its hyperparameters as: initial learning rate lr = 0.05, momen = 0.9, eps = 1e-8, weight\_decay = 5e-4. The warm-up and linear decay strategies are used to adjust the learning rate. The batch size is set to \Rev{32~(ResNet), 10~(PVTv2/CycleMLP) or 8~(VGG/Swin),} and the maximum number of epochs is set to 60~(\Rev{ResNet-based training} takes $\sim$2.5 hours). In addition, we use apex\footnote{\Rev{~\url{https://github.com/NVIDIA/apex}}} and fp16 to accelerate the training process. Gradient clipping is also used to prevent gradient explosion. The inference process \Rev{of the ResNet-based architecture} for a 352$\times$352 image only takes \Rev{0.0164s}, including the IO time.

\subsection{Evaluation Metrics}
We use five metrics to evaluate our model and the existing state-of-the-art algorithms:

\textbf{(1) MAE \Rev{($M$)}} evaluates the average 
pixel-wise difference between the predicted saliency map ($\textit{P}$) and the ground-truth map ($\textit{G}$). We normalize $\textit{P}$ and $\textit{G}$ to $[0,1]$, so the MAE score can be computed as $M =\frac{1}{ \textit{W} \times \textit{H} } \sum_{\textit{x}=1}^{\textit{W}} \sum_{\textit{y}=1}^{\textit{H}}|P(\textit{x},\textit{y})-G(\textit{x}, \textit{y})|.$ 


\textbf{(2) Weighted F-measure~($F_{\beta}^{\omega}$)}~\cite{margolin2014evaluate}  offers an intuitive generalization of
$F_{\beta}$, and is defined as $F_{\beta}^{\omega}=\frac{\left(1+\beta^{2}\right) \text{Precision}^{\omega} \cdot \text {Recall}^{\omega}}{\beta^{2} \cdot \text {Precision}^{\omega}+\text {Recall}^{\omega}}$. As a \Rev{widely} adopted metric \cite{li2016deep,li2018contour,liu2019employing,zhang2020multistage,wang2018salient2,fan2018SOC,piao2019depth,li2017instance,zeng2019towards,zhang2017amulet}, $F_{\beta}^{\omega}$ can handle the interpolation, dependency and equal-importance issues, which might cause inaccurate evaluation by MAE and F-measure\Rev{~\cite{achanta2009frequency}}. As suggested in~\cite{borji2015salient}, we set $\beta^{2}$ to 1.0 to emphasize the precision over recall. By assigning different weights~($\omega$) to different errors based on the specific location and neighborhood information, $F_{\beta}^{\omega}$ extends \Rev{the F-measure} to non-binary \Rev{evaluation}.

\textbf{(3) S-measure~($S_{m}$)}~\cite{cheng2021structure} focuses on evaluating the structural similarity, which is much closer to human visual perception. It is computed as $S_m=m \mathrm{s}_{o}+(1-m) \mathrm{s}_{r}$, where $\mathrm{s}_{o}$ and $\mathrm{s}_{r}$ denote the object-aware and region-aware structural similarity and $m$ is set to 0.5, as suggested in \cite{cheng2021structure}.

\textbf{(4) E-measure~($E_\xi$)}~\cite{fan2018enhanced} combines the local pixel values with the image-level mean value in one term \Rev{and can be computed as:} $E_\xi=\frac{1}{W \times H} \sum_{x=1}^{W} \sum_{y=1}^{H} \theta\left(\xi\right)$, where $\xi$ is the alignment matrix and $\theta\left(\xi\right)$ indicates the enhanced alignment matrix. We adopt mean E-measure ($E_\xi^{m}$) as our final evaluation metric.

\textbf{(5) FNR} is the false negative ratio, which is computed by:
\begin{footnotesize}
\begin{equation}
\label{fn}
\Rev{FN(x, y)=\left\{
\begin{aligned}
1 &, G(x,y) = 1~\&~ P(x, y) = 0,\\
0 &, ~others.
\end{aligned}
\right.}
\end{equation}
\end{footnotesize}

\begin{footnotesize}
 \begin{equation} 
    \Rev{FNR = \frac{\sum_{x=1}^{W} \sum_{y=1}^{H} FN\left(x, y\right)}{\sum_{x=1}^{W} \sum_{y=1}^{H} G\left(x, y\right)} \times 100\%,}
    \label{equ:eval5}
\end{equation}   
\end{footnotesize}
\vspace{-5pt}
\\
where $FN$ is the pixel-level indicator that determines whether a pixel is a false negative. We show several examples of FNR in \figref{fig:fnr2}. It clearly and accurately reflects the \textit{integrity} of prediction results and is sensitive at the macro and micro level.

\subsection{Comparison with the SOTAs}
We compare the proposed approach with \Red{14} very recent state-of-the-art methods, including 
\Red{Condinst~\cite{tian2020conditional}, PointRend~\cite{kirillov2020pointrend},}
PiCANet~\cite{liu2018picanet},
RAS~\cite{chen2018reverse}, 
AFNet~\cite{feng2019attentive}, BASNet~\cite{qin2019basnet}, CPD~\cite{wu2019cascaded},  
EGNet~\cite{zhao2019egnet}, SCRN~\cite{wu2019stacked}, F3Net~\cite{F3Net},  MINet~\cite{pang2020multi}, ITSD~\cite{zhou2020interactive}, GateNet~\cite{GateNet}
\Rev{and VST~\cite{liu2021visual}}.

\begin{figure*}[t!]
	\centering
	\begin{overpic}[width=.98\textwidth]{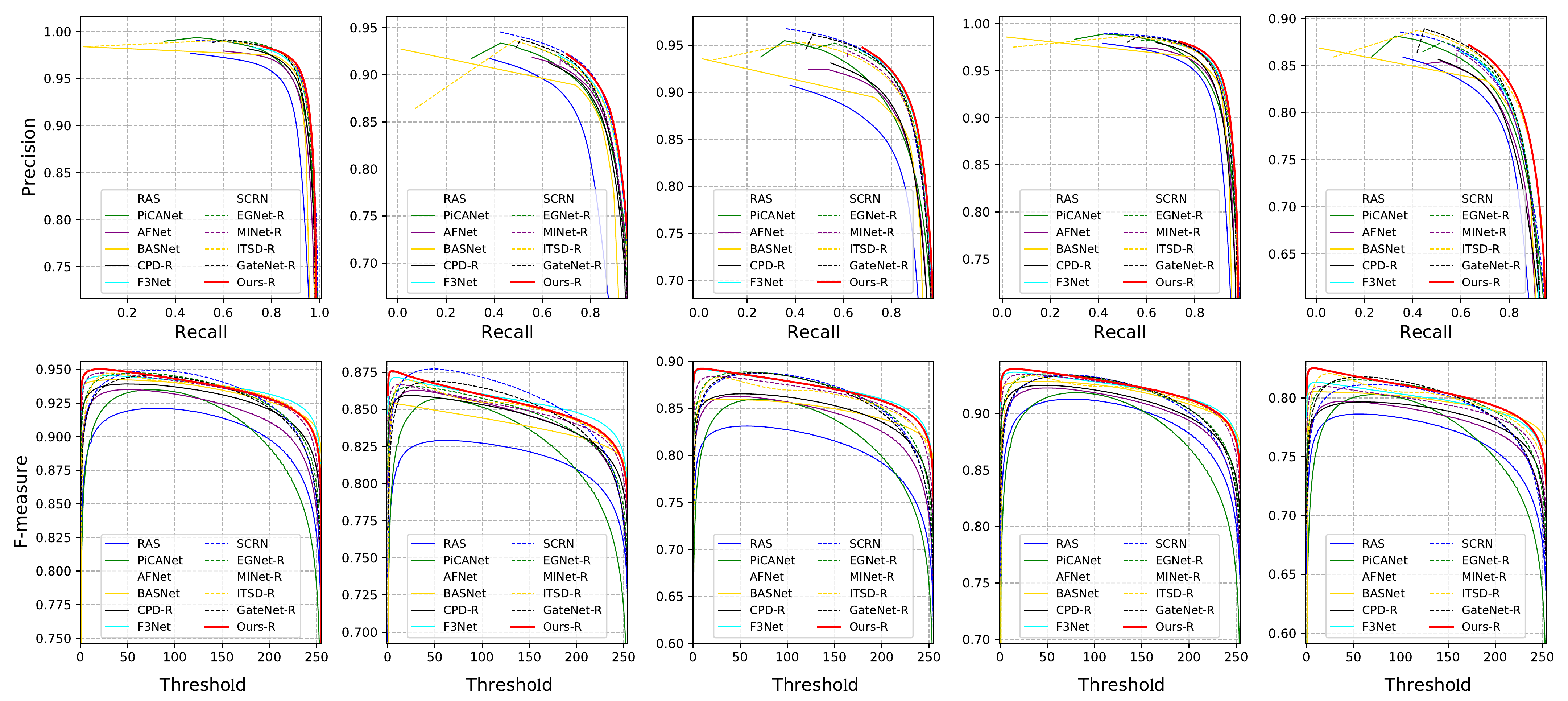} 
	\put(9.7,36.8){\normalsize{\textbf{ECSSD}}}
	\put(9.7,14.3){\normalsize{\textbf{ECSSD}}}
	\put(27.9,36.8){\normalsize{\textbf{PASCAL-S}}}
	\put(27.9,14.3){\normalsize{\textbf{PASCAL-S}}}
	\put(49.1,36.8){\normalsize{\textbf{DUTS}}}
	\put(49.1,14.3){\normalsize{\textbf{DUTS}}}
	\put(67.9,36.8){\normalsize{\textbf{HKU-IS}}}
	\put(67.9,14.3){\normalsize{\textbf{HKU-IS}}}
	\put(87.1,36.8){\normalsize{\textbf{OMRON}}}
	\put(87.1,14.3){\normalsize{\textbf{OMRON}}}
	\end{overpic}
	\caption{{Precision-recall and F-measure curves of the proposed method and other SOTA algorithms on five popular SOD datasets.} }
\label{fig:curves}
\end{figure*}

\begin{figure*}[t!]
	\centering
	\begin{overpic}[width=.98\textwidth]{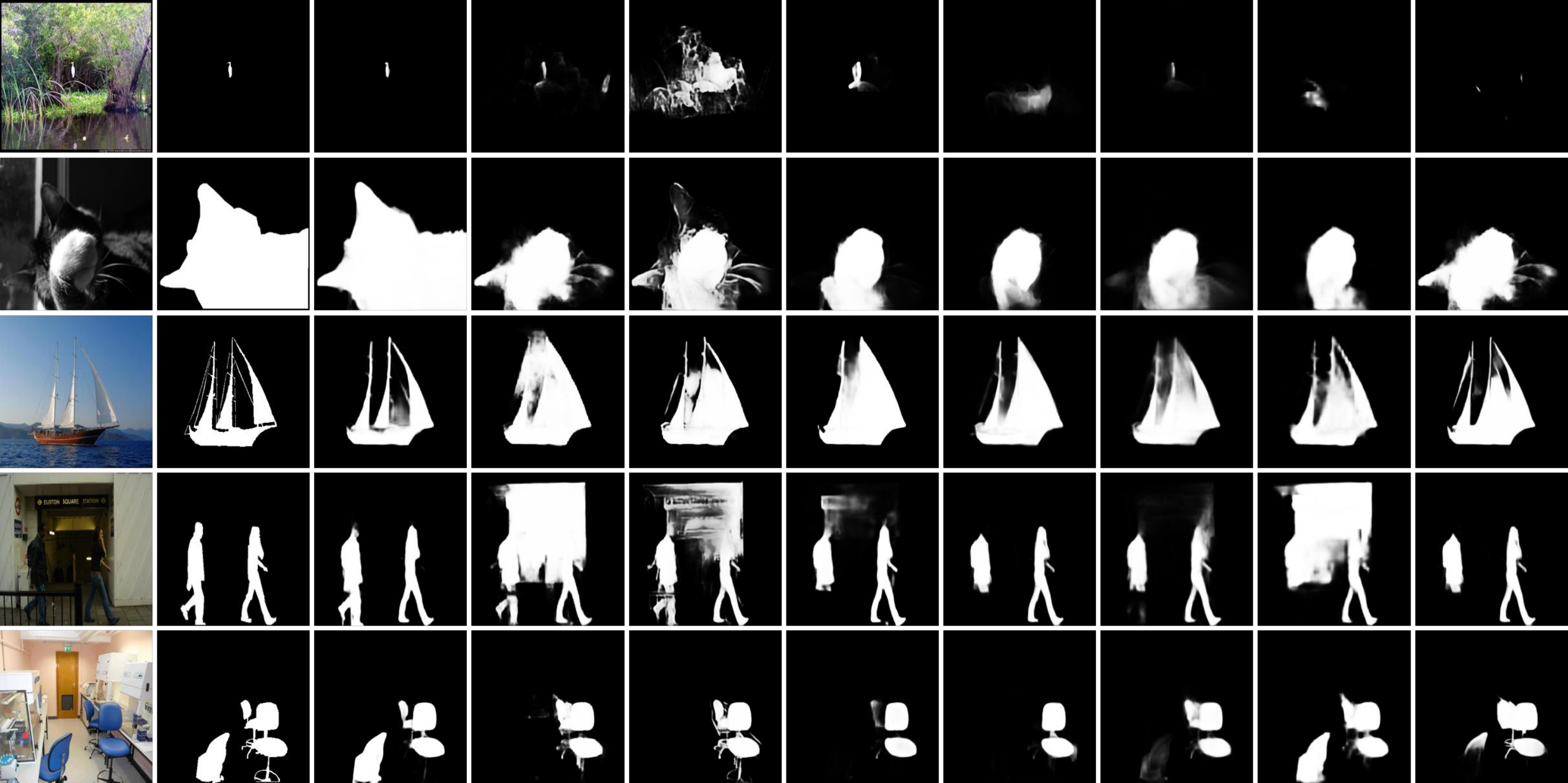}
	\put(3.1,-2){\footnotesize{Image}}
	\put(14.3,-2){\footnotesize{GT}}
	\put(22.8,-2){\footnotesize{Ours-R}}
	\put(32.1,-2){\footnotesize{MINet-R}}
	\put(42.8,-2){\footnotesize{ITSD-R}}
	\put(53,-2){\footnotesize{F3Net}}
	\put(62,-2){\footnotesize{EGNet-R}}
	\put(73.3,-2){\footnotesize{SCRN}}
	\put(83,-2){\footnotesize{CPD-R}}
	\put(92.3,-2){\footnotesize{BASNet}}
	
	\put(33.2,-4){\footnotesize{ ~\cite{pang2020multi}}}
	\put(43.2,-4){\footnotesize{ ~\cite{zhou2020interactive}}}	
	\put(53.5,-4){\footnotesize{~\cite{F3Net}}}
	\put(63.5,-4){\footnotesize{~\cite{zhao2019egnet}}}
	\put(73.8,-4){\footnotesize{~\cite{wu2019stacked}}}
	\put(83.5,-4){\footnotesize{~\cite{wu2019cascaded}}}
	\put(93.2,-4){\footnotesize{~\cite{qin2019basnet}}}
	\end{overpic}
    \vspace*{10pt}
	\caption{Qualitative comparison of our method with \Rev{seven} SOTA methods.  Unlike other baseline methods, our method not only accurately locates the salient object but also produces sharper edges with fewer background distractors for various scenes.}
	\label{fig:vc}
\end{figure*}
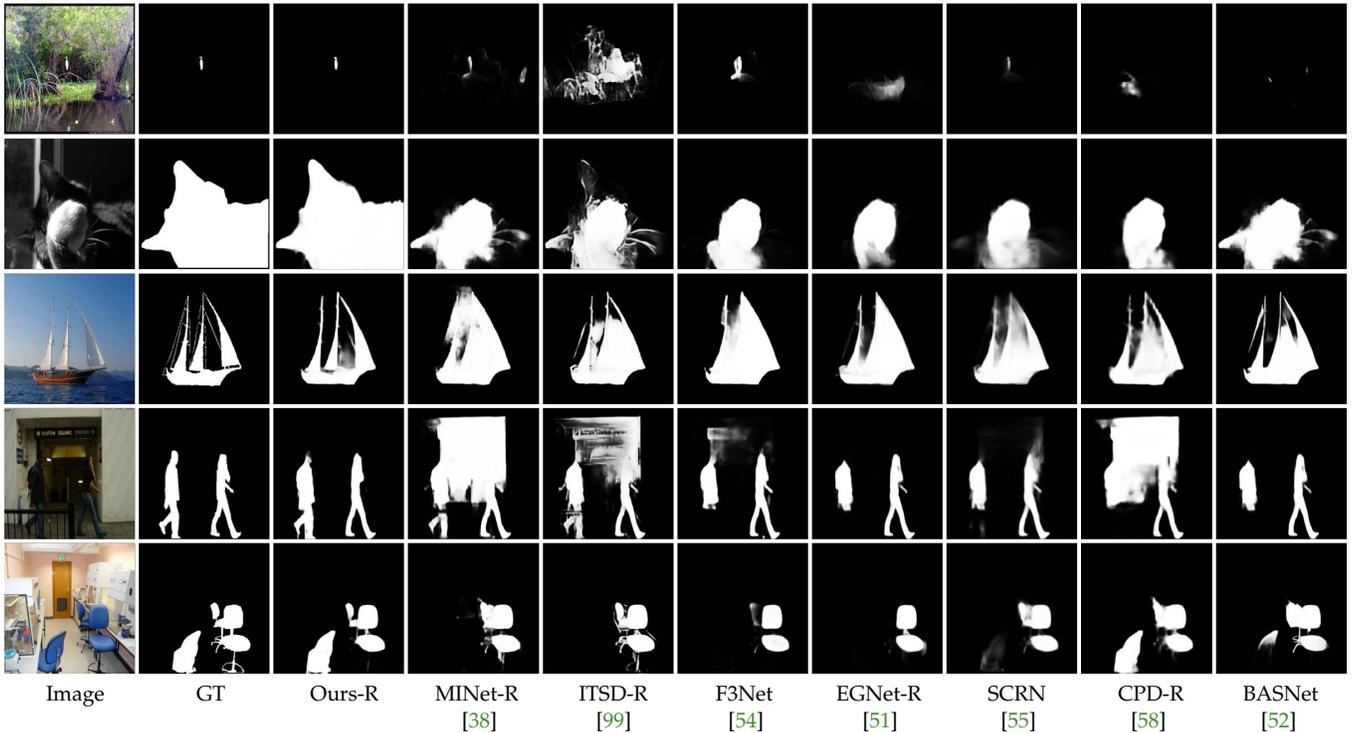 

\begin{table*}[t!]
	\caption{Comparison between the proposed method and other SOTA methods on the SOC test set. For ``$\uparrow$'' and ``$\downarrow$'', higher and lower scores indicate better results, respectively. 
	In the last column, $\dagger$ indicates that the parameters are trained on the SOC training set.}
	\centering
	\footnotesize
    \renewcommand{\arraystretch}{1.0}
    \setlength\tabcolsep{0.3pt}
	\begin{tabular}{l|r|cccccccccccccccc|c}
		\toprule
		\textbf{Attr}	&	\textbf{Metrics}	&	\tabincell{c}{\textbf{Amulet}\\\cite{zhang2017amulet}}	&	\tabincell{c}{\textbf{~~DSS~~}\\\cite{hou2019deeply}}	&	\tabincell{c}{\textbf{~NLDF~}\\\cite{luo2017non}}	&	\tabincell{c}{\textbf{C2SNet}\\\cite{li2018contour}}	&	\tabincell{c}{\textbf{~~SRM~~}\\\cite{DBLP:conf/iccv/WangBZZL17}}	&	\tabincell{c}{\textbf{R3Net}\\\cite{deng2018r3net}}	&	\tabincell{c}{\textbf{~BMPM~}\\\cite{zhang2018bi}}	&	\tabincell{c}{\textbf{~DGRL~}\\\cite{wang2018detect}}	&	\tabincell{c}{\textbf{PiCA-R}\\\cite{liu2018picanet}}	&	\tabincell{c}{\textbf{RANet}\\\cite{chen2020reverse}}	&	\tabincell{c}{\textbf{AFNet}\\\cite{feng2019attentive}}	&	\tabincell{c}{\textbf{~~CPD~~}\\\cite{wu2019cascaded}}	&	\tabincell{c}{\textbf{PoolNet}\\\cite{liu2019simple}}	&	\tabincell{c}{\textbf{EGNet}\\\cite{zhao2019egnet}}	&	\tabincell{c}{\textbf{~BANet~}\\\cite{su2019selectivity}}	&	\tabincell{c}{\textbf{~SCRN~}\\\cite{wu2019stacked}}	&	\tabincell{c}{\textbf{ICON-R}\\(Ours)} 	\\
		\toprule
\multirow{4}{*}{\textbf{AC}}
	&	$S_m$~$\uparrow$	&	0.752	&	0.753	&	0.737	&	0.755	&	0.791	&	0.713	&	0.780	&	0.790	&	0.792	&	0.708	&	0.796	&	0.799	&	0.795	&	0.806	&	0.806	&	0.809	& \textbf{0.835}/\textbf{0.835}$\dagger$	
	\\
	&	$E_\xi^{m}$~$\uparrow$	
	&	0.791	&	0.788	&	0.784	&	0.807	&	0.824	&	0.753	&	0.815	&0.853	&	0.815	&0.765	&	0.852	&	0.843	&	0.846	&	0.854	&	0.858	&	0.849	&	\textbf{0.891}/{0.889}$\dagger$	\\
    &	$F^w_\beta$~$\uparrow$	&	0.620	&	0.629	&	0.620	&	0.647	&	0.690	&	0.593	&	0.680	&	0.718	&	0.682	&	0.603	&	0.712	&	0.727	&	0.713	&	0.731	&	0.740	&	0.724	& \textbf{0.784}/\textbf{0.784}$\dagger$
    \\
	&	$M\downarrow$	&	0.120	&	0.113	&	0.119	&	0.109	&	0.096	&	0.135	&	0.098	&	0.081	&	0.093	&	0.132	&	0.084	&	0.083	&	0.094	&	0.085	&	0.086	&	0.078	& \textbf{0.062}/{0.064}$\dagger$
	\\

	\midrule
\multirow{4}{*}{\textbf{BO}}
	&	$S_m\uparrow$	&	0.574	&	0.561	&	0.568	&	0.654	&	0.614	&	0.437	&	0.604	&	0.684	&	\textbf{0.729}	&	0.421	&	0.658	&	0.647	&	0.561	&	0.528	&	0.645	&	0.698	&0.714/0.713$\dagger$	\\
	&	$E_\xi^{m}$~$\uparrow$		&	0.551	&	0.537	&	0.539	&	0.661	&	0.616	&	0.419	&	0.620	&	0.725	&	0.741	&	0.404	&	0.698	&	0.665	&	0.554	&	0.528	&	0.650	&	0.706	& {0.740}/\textbf{0.743}$\dagger$
	\\
    &	$F^w_\beta$~$\uparrow$	&	0.612	&	0.614	&	0.622	&	0.730	&	0.667	&	0.456	&	0.670	&	0.786	&\textbf{0.799}		&	0.453	&	0.741	&	0.739	&	0.610	&	0.585	&	0.720	&	0.778	& {0.794}/0.794$\dagger$
    \\
	&	$M\downarrow$	&	0.346	&	0.356	&	0.354	&	0.267	&	0.306	&	0.445	&	0.303	&0.215	&	{0.200}	&	0.454	&	0.245	&	0.257	&	0.353	&	0.373	&	0.271	&	0.224	& {0.200}/\textbf{0.199}$\dagger$	\\

	\midrule
\multirow{4}{*}{\textbf{CL}}
	&	$S_m\uparrow$	&	0.763	&	0.722	&	0.713	&	0.742	&	0.759	&	0.659	&	0.761	&	0.770	&	0.787	&	0.624	&	0.768	&	0.773	&	0.760	&	0.757	&	0.784	&	0.795	&{0.789}/\textbf{0.802}$\dagger$		\\
	&	$E_\xi^{m}$~$\uparrow$	&	0.789	&	0.763	&	0.764	&	0.789	&	0.793	&	0.710	&	0.801	&	0.824	&	0.794	&	0.715	&	0.802	&	0.821	&	0.801	&	0.790	&	0.824	&	0.820	&{0.829}/\textbf{0.855}$\dagger$		\\
    &	$F^w_\beta$~$\uparrow$	&	0.663	&	0.617	&	0.614	&	0.655	&	0.665	&	0.546	&	0.678	&	0.714	&	0.692	&	0.542	&	0.696	&	0.724	&	0.681	&	0.677	&	0.726	&	0.717	&{0.732}/\textbf{0.754}$\dagger$		\\
	&	$M\downarrow$	&	0.141	&	0.153	&	0.159	&	0.144	&	0.134	&	0.182	&	0.123	&	0.119	&	0.123	&	0.188	&	0.119	&	0.114	&	0.134	&	0.139	&	0.117	&	{0.113}	& 0.113/\textbf{0.102}$\dagger$
	\\
	\midrule
\multirow{4}{*}{\textbf{HO}}
	&	$S_m\uparrow$	&	0.791	&	0.767	&	0.755	&	0.768	&	0.794	&	0.740	&	0.781	&	0.791	&	0.809	&	0.713	&	0.798	&	0.803	&	0.815	&	0.802	&	0.819	&	{0.823}	& {0.818}/\textbf{0.830}$\dagger$
	\\
	&	$E_\xi^{m}$~$\uparrow$	&	0.810	&	0.796	&	0.798	&	0.805	&	0.819	&	0.782	&	0.813	&	0.833	&	0.819	&	0.777	&	0.834	&	0.845	&	0.846	&	0.829	&	0.850	&	0.842	& {0.852}/\textbf{0.865}$\dagger$
	\\
    &	$F^w_\beta$~$\uparrow$	&	0.688	&	0.660	&	0.661	&	0.668	&	0.696	&	0.633	&	0.684	&	0.722	&	0.704	&	0.626	&	0.722	&0.751	&	0.739	&	0.720	&	0.754	&	0.743	& {0.752}/\textbf{0.771}$\dagger$
    \\
	&	$M\downarrow$	&	0.119	&	0.124	&	0.126	&	0.123	&	0.115	&	0.136	&	0.116	&	0.104	&	0.108	&	0.143	&	0.103	&	0.097	&	0.100	&	0.106	&	0.094	&	0.096	&{0.092}/\textbf{0.087}$\dagger$		\\
	\midrule
\multirow{4}{*}{\textbf{MB}}
	&	$S_m\uparrow$	&	0.712	&	0.719	&	0.685	&	0.720	&	0.742	&	0.657	&	0.762	&	0.744	&	0.775	&	0.696	&	0.734	&	0.754	&0.751	&	0.762	&	0.764	&{0.792}	&0.774/\textbf{0.821}$\dagger$	
	\\
	&	$E_\xi^{m}$~$\uparrow$	&	0.739	&	0.753	&	0.740	&	0.778	&	0.778	&	0.697	&	0.812	&	{0.823}	&	0.813	&	0.761	&	0.762	&	0.804	&	0.779	&	0.789	&	0.803	&	0.817	& 0.828/\textbf{0.866}$\dagger$
	\\
    &	$F^w_\beta$~$\uparrow$	&	0.561	&	0.577	&	0.551	&	0.593	&	0.619	&	0.489	&	0.651	&	0.655	&	0.637	&	0.576	&	0.626	&	0.679	&	0.642	&	0.649	&	0.672	&	{0.690}	&0.699/\textbf{0.768}$\dagger$
    \\
	&	$M\downarrow$	&	0.142	&	0.132	&	0.138	&	0.128	&	0.115	&	0.160	&	0.105	&	0.113	&{0.099}		&	0.139	&	0.111	&	0.106	&	0.121	&	0.109	&	0.104	&	0.100	&0.100/\textbf{0.076}$\dagger$		\\
	\midrule
\multirow{4}{*}{\textbf{OC}}
	&	$S_m\uparrow$	&	0.735	&	0.718	&	0.709	&	0.738	&	0.749	&	0.653	&	0.752	&	0.747	&	0.765	&	0.641	&	0.771	&	0.750	&	0.756	&	0.754	&	0.765	&	{0.775}	&0.771/\textbf{0.791}$\dagger$	\\
	&	$E_\xi^{m}$~$\uparrow$	&	0.763	&	0.760	&	0.755	&	0.784	&	0.780	&	0.706	&	0.800	&	0.808	&	0.784	&	0.718	&	{0.820}	&	0.810	&	0.801	&	0.798	&	0.809	&	0.800	&0.817/\textbf{0.831}$\dagger$		\\
    &	$F^w_\beta$~$\uparrow$	&	0.607	&	0.595	&	0.593	&	0.622	&	0.630	&	0.520	&	0.644	&	0.659	&	0.638	&	0.527	&	{0.680}	&	0.672	&	0.659	&	0.658	&	0.678	&	0.673	&0.683/\textbf{0.710}$\dagger$		\\
	&	$M\downarrow$	&	0.143	&	0.144	&	0.149	&	0.130	&	0.129	&	0.168	&	0.119	&	0.116	&	0.119	&	0.169	&{0.109}	&	0.115	&	0.119	&	0.121	&	0.112	&	0.111	&{0.106}/\textbf{0.100}$\dagger$		\\
	\midrule
\multirow{4}{*}{\textbf{OV}}
	&	$S_m\uparrow$	&	0.721	&	0.700	&	0.688	&	0.728	&	0.745	&	0.624	&	0.751	&	0.762	&	{0.781}	&	0.611	&	0.761	&	0.748	&	0.747	&	0.752	&	0.779	&	0.774	&0.779/\textbf{0.802}$\dagger$		\\
	&	$E_\xi^{m}$~$\uparrow$	&	0.751	&	0.737	&	0.736	&	0.790	&	0.779	&	0.663	&	0.807	&	0.828	&	0.810	&	0.664	&	0.817	&	0.803	&	0.795	&	0.802	&	{0.835}	&	0.808	&0.834/\textbf{0.846}$\dagger$	\\
    &	$F^w_\beta$~$\uparrow$	&	0.637	&	0.622	&	0.616	&	0.671	&	0.682	&	0.527	&	0.701	&	0.733	&	0.721	&	0.529	&	0.723	&	0.721	&	0.697	&	0.707	&	{0.752}	&	0.723	&0.749/\textbf{0.768}$\dagger$	\\
	&	$M\downarrow$	&	0.173	&	0.180	&	0.184	&	0.159	&	0.150	&	0.216	&	0.136	&	0.125	&	0.127	&	0.217	&	0.129	&	0.134	&0.148	&	0.146	&	{0.119}	&	0.126	&0.120/\textbf{0.108}$\dagger$		\\

	\midrule
\multirow{4}{*}{\textbf{SC}}
	&	$S_m\uparrow$	&	0.768	&	0.761	&	0.745	&	0.756	&	0.783	&	0.716	&	0.799	&	0.772	&	0.784	&	0.724	&	0.808	&	0.793	&	0.807	&	0.793	&	0.807	&	{0.809}	&0.803/\textbf{0.824}$\dagger$	\\
	&	$E_\xi^{m}$~$\uparrow$	&	0.794	&	0.799	&	0.788	&	0.806	&	0.814	&	0.765	&	0.841	&	0.837	&	0.799	&	0.792	&	0.854	&	{0.858}	&	0.856	&	0.844	&	0.851	&	0.843	&0.860/\textbf{0.882}$\dagger$		\\
    &	$F^w_\beta$~$\uparrow$	&	0.608	&	0.599	&	0.593	&	0.611	&	0.638	&	0.550	&	0.677	&	0.669	&	0.627	&	0.594	&	0.696	&	{0.708}	&	0.695	&	0.678	&	0.706	&0.691	&{0.714}/\textbf{0.745}$\dagger$		\\
	&	$M\downarrow$	&	0.098	&	0.098	&	0.101	&	0.100	&	0.090	&	0.114	&	0.081	&	0.087	&	0.093	&	0.110	&	0.076	&	0.080	&	{0.075}	&	0.083	&	0.078	&	0.078	&0.080/\textbf{0.073}$\dagger$	 \\
	\midrule
\multirow{4}{*}{\textbf{SO}}
	&	$S_m\uparrow$	&	0.718	&	0.713	&	0.703	&	0.706	&	0.737	&	0.682	&	0.732	&	0.736	&	0.748	&	0.682	&	0.746	&	0.745	&	{0.768}	&	0.749	&	0.755	&	0.767	&0.763/\textbf{0.801}$\dagger$		\\
	&	$E_\xi^{m}$~$\uparrow$	&	0.745	&	0.756	&	0.747	&	0.752	&	0.769	&	0.732	&	0.780	&	0.802	&	0.766	&	0.759	&	0.792	&	0.804	&	{0.814}	&	0.784	&	0.801	&	0.797	&0.816/\textbf{0.848}$\dagger$	\\
    &	$F^w_\beta$~$\uparrow$	&	0.523	&	0.524	&	0.526	&	0.531	&	0.561	&	0.487	&	0.567	&	0.602	&	0.566	&	0.518	&	0.596	&	0.623	&	0.626	&	0.594	&	0.621	&	0.614	&{0.634}/\textbf{0.689}$\dagger$	\\
	&	$M\downarrow$	&	0.119	&	0.109	&	0.115	&	0.116	&	0.099	&	0.118	&	0.096	&	0.092	&	0.095	&	0.113	&	0.089	&	0.091	&	0.087	&	0.098	&	0.090	&	{0.082}	&0.087/\textbf{0.073}$\dagger$	\\
	\bottomrule
	\end{tabular}
	\label{tab:compSOC}
\end{table*}

\subsubsection{Quantitative Evaluation}
\tabref{tab:comp1} reports the quantitative results on six traditional benchmark datasets, in which we compare our method with the \Red{14} state-of-the-art algorithms in terms of S-measure, E-measure, weighted F-measure, and MAE. Our model is clearly better than the other baseline methods. Besides, we also show the FNR results of ours and the baseline methods in \Rev{\figref{fig:fnr1}}. As can be seen, our approach achieves the lowest FNR scores across all datasets. Visual comparison (see \figref{fig:fnr2}) also demonstrates the efficiency of our method for capturing integral objects. In fact, our \ourmodel~performs favorably against the existing methods across all datasets in terms of nearly all evaluation metrics. This demonstrates its strong capability in dealing with challenging inputs. In addition, we present the precision-recall~\cite{cheng2014global} and F-measure curves~\cite{achanta2009frequency} in \figref{fig:curves}. The solid red lines belonging to the proposed method are obviously higher than the other curves, which further demonstrates the effectiveness of the proposed method based on integrity learning.

\subsubsection{Visual Comparison}
\figref{fig:vc} provides visual comparison between our approach and the baseline methods. As can be observed, our ICON generates more accurate saliency maps for various challenging cases, \textit{e.g.}, \Rev{small objects (1$^{st}$), large objects (2$^{nd}$ row),  delicate structures (3$^{rd}$ row), low-contrast (4$^{th}$ row), and multiple objects (5$^{th}$ row).} Besides, our framework can detect salient targets integrally and noiselessly. The above results demonstrate the accuracy and robustness of the proposed method.

\subsubsection{Attribute-Based Analysis}
In addition to the most frequently used saliency detection datasets, we also evaluate our method on another challenging SOC dataset~\cite{Fan2021SOC,fan2018SOC}. When compared with the previous six SOD datasets, this dataset contains many more complicated scenes. In addition, the SOC dataset categorizes images according to nine different attributes, including AC (appearance change), BO (big object), CL (clutter), HO (heterogeneous object), MB (motion blur), OC (occlusion), OV (out-of-view), SC (shape complexity), and SO (small object). 

In \tabref{tab:compSOC}, we compare our \ourmodel~with 16 state-of-the-art methods, including Amulet \cite{zhang2017amulet}, DSS \cite{hou2019deeply}, NLDF \cite{luo2017non}, C2SNet \cite{li2018contour}, SRM \cite{DBLP:conf/iccv/WangBZZL17}, R3Net \cite{deng2018r3net}, BMPM \cite{zhang2018bi}, DGRL \cite{wang2018detect}, PiCANet-R (PiCA-R) \cite{liu2020picanet}, RANet \cite{chen2020reverse}, AFNet \cite{feng2019attentive}, CPD \cite{wu2019cascaded}, PoolNet \cite{liu2019simple}, EGNet \cite{zhao2019egnet}, BANet \cite{su2019selectivity} and SCRN \cite{wu2019stacked} in terms of attribute-based performance. 
As seen, our \ourmodel~achieves clear performance improvement over the existing methods. 

\begin{table*}[t!]
	\footnotesize
	\centering
	\renewcommand{\arraystretch}{1}
	\setlength\tabcolsep{7.88pt}
	\caption{Ablation analysis of our baseline gradually including the newly proposed components. The best results are shown in \textbf{bold}.}
	\label{tab:Baseline}
	\begin{tabular}{l|l|cccc|cccc|cccc}
		\toprule
		\multirow{2}{*}{ID} & \multirow{2}{*}{Component Settings} & \multicolumn{4}{c|}{OMRON~\cite{yang2013saliency}} & \multicolumn{4}{c|}{HKU-IS \cite{li2015visual}} & \multicolumn{4}{c}{DUTS-TE~\cite{wang2017learning}} \\ 
		&&  $S_m$$\uparrow$    & $E_\xi^{m}$$\uparrow$       & $F_\beta^w$$\uparrow$    &   $M$$\downarrow$     &  $S_m$$\uparrow$    & $E_\xi^{m}$$\uparrow$       & $F_\beta^w$$\uparrow$    &   $M$$\downarrow$    &  $S_m$$\uparrow$    & $E_\xi^{m}$$\uparrow$       & $F_\beta^w$$\uparrow$    &   $M$$\downarrow$
		\\ \hline
		1 & Baseline &
		0.832 & 0.854  & 0.731 & 0.064    & 0.902 & 0.930  & 0.866  & 0.043    & 0.861  & 0.879 & 0.801  & 0.049
		\\
		2 & +DFA &
		0.837  & 0.857  & 0.740 & 0.063     & 0.913 & 0.939  & 0.875  & 0.035  & 0.879  & 0.886  & 0.818  & 0.046
		\\
		3 & +DFA+ICE 
		& 0.840 &0.869 & 0.753 & 0.059   & 0.918 & 0.951 & 0.895 & 0.031  & 0.887 & 0.916 & 0.825 & 0.038
		\\
		\rowcolor{mygray}
		4 & +DFA+ICE+PWV 
		& \textbf{0.844} & \textbf{0.876}  & \textbf{0.761}  & \textbf{0.057}     & \textbf{0.920} & \textbf{0.953}  & \textbf{0.902}  & \textbf{0.029}     & \textbf{0.888}  & \textbf{0.924}  & \textbf{0.836}  & \textbf{0.037}	                
		\\ \bottomrule
	\end{tabular}
\end{table*}

\begin{table*}[t!]
	\footnotesize
	\centering
	\renewcommand{\arraystretch}{1}
	\setlength\tabcolsep{7.38pt}
	\caption{\Rev{Ablation analysis of different feature enhancement methods (FEMs) compared with the DFA module of our ICON. }}
	\label{tab:FEMs}
	\begin{tabular}{l|l|cccc|cccc|cccc}
		\toprule
		\multirow{2}{*}{ID} & \multirow{2}{*}{FEMs Settings} & \multicolumn{4}{c|}{OMRON~\cite{yang2013saliency}} & \multicolumn{4}{c|}{HKU-IS \cite{li2015visual}} & \multicolumn{4}{c}{DUTS-TE~\cite{wang2017learning}} \\
		&&  $S_m$$\uparrow$    & $E_\xi^{m}$$\uparrow$       & $F_\beta^w$$\uparrow$    &   $M$$\downarrow$     &  $S_m$$\uparrow$    & $E_\xi^{m}$$\uparrow$       & $F_\beta^w$$\uparrow$    &   $M$$\downarrow$    &  $S_m$$\uparrow$    & $E_\xi^{m}$$\uparrow$       & $F_\beta^w$$\uparrow$    &   $M$$\downarrow$
		\\ \hline
		\rowcolor{mygray}
		2 & +DFA &
		0.837  & \textbf{0.857}  & \textbf{0.740} & 0.063     & \textbf{0.913} & 0.939  & \textbf{0.875}  & 0.035  & \textbf{0.879}  & \textbf{0.886}  & \textbf{0.818}  & 0.046
		\\
		5 & \multicolumn{1}{l|}{+Inception\cite{szegedy2016rethinking}}   &
		0.839  & 0.853  & \textbf{0.740}  & 0.064     & 0.909 & 0.936  & 0.869  & 0.037     & 0.875  & \textbf{0.886}  & 0.817  & \textbf{0.043}
		\\
		6 & \multicolumn{1}{l|}{+ASPP\cite{chen2017deeplab}}   &
		\textbf{0.840}  & 0.855  & 0.738  & \textbf{0.061}     & 0.912 & \textbf{0.943}  & 0.869  & \textbf{0.034}     & 0.877  & 0.882  & \textbf{0.818}  & \textbf{0.043}
		\\
		7 & \multicolumn{1}{l|}{+PSP\cite{zhao2017pyramid}}   &
		0.835  & 0.851  & 0.738  & 0.063     & 0.906 & 0.935  & 0.870  & 0.036     & 0.878  & 0.884  & 0.816  & 0.045
		\\ 
		8 & \multicolumn{1}{l|}{+DFA~(3xOriConv)}   &
		0.833  & 0.855  & 0.733  & 0.062    & 0.909 & 0.938  & 0.868  & 0.035     & 0.874  & 0.875  & 0.810  & 0.046
		\\ 				
		9 & \multicolumn{1}{l|}{+DFA~(3xAtrConv~\cite{chen2017deeplab})}   
		&  0.830 & 0.849  & 0.729  &  0.066    & 0.906 & 0.933  &  0.869 &   0.038  & 0.872  & 0.880 &  0.811 & 0.047				
		\\ 					
		10 & \multicolumn{1}{l|}{+DFA~(3xAsyConv~\cite{ding2019acnet})}   &
		0.837  & 0.854  & 0.737  & 0.064     & 0.909 & 0.937  & 0.873  & 0.035     & \textbf{0.879}  & 0.885  & 0.817  & 0.044	
		\\
		\bottomrule
	\end{tabular}
\end{table*}

\begin{table*} [t!]
	\footnotesize
	\centering
	\renewcommand{\arraystretch}{1.0}
	\setlength\tabcolsep{8.07pt}
	\caption{Ablation analysis of our ICE and other alternative methods  using various attention mechanisms. 
	}
	\label{tab:ICE}
	
	\begin{tabular}{l|l|cccc|cccc|cccc}
		\toprule
		\multirow{2}{*}{ID} & \multirow{2}{*}{Attention Settings} & \multicolumn{4}{c|}{OMRON~\cite{yang2013saliency}} & \multicolumn{4}{c|}{HKU-IS \cite{li2015visual}} & \multicolumn{4}{c}{DUTS-TE~\cite{wang2017learning}} \\ 
		&&  $S_m$$\uparrow$    & $E_\xi^{m}$$\uparrow$       & $F_\beta^w$$\uparrow$    &   $M$$\downarrow$     &  $S_m$$\uparrow$    & $E_\xi^{m}$$\uparrow$       & $F_\beta^w$$\uparrow$    &   $M$$\downarrow$    &  $S_m$$\uparrow$    & $E_\xi^{m}$$\uparrow$       & $F_\beta^w$$\uparrow$    &   $M$$\downarrow$
		\\ \hline
		\rowcolor{mygray}
		3 & +DFA+ICE 
		& 0.840 &\textbf{0.869 }& \textbf{0.753} & 0.059   & \textbf{0.918} & \textbf{0.951} & \textbf{0.895} & \textbf{0.031}  & 0.887 & \textbf{0.916} & 0.825 & \textbf{0.038}
		\\
		11 & +DFA+SE\cite{hu2020squeeze} &
		0.839  & 0.861  & 0.720  & 0.061    & 0.909 & 0.943 & 0.877  & 0.034     & \textbf{0.888}  & 0.895 & 0.831  & 0.039	
		\\
		12 & +DFA+CBAM\cite{woo2018cbam} &
		\textbf{0.842} & 0.864  & 0.739  & \textbf{0.058}     & 0.917 & 0.946  & 0.891  & \textbf{0.031}     & 0.885  & 0.907  & \textbf{0.832}  & 0.039
		\\
		13 & +DFA+GCT\cite{yang2020gated} &
		0.838  & 0.857  & 0.712  & 0.062     & 0.901 & 0.937  & 0.874  & 0.033     & 0.883  & 0.905  & 0.821  & 0.041	
		\\ \bottomrule
	\end{tabular}
\end{table*}

\begin{figure}[h!]
	\centering
	\begin{overpic}[width=0.47\textwidth]{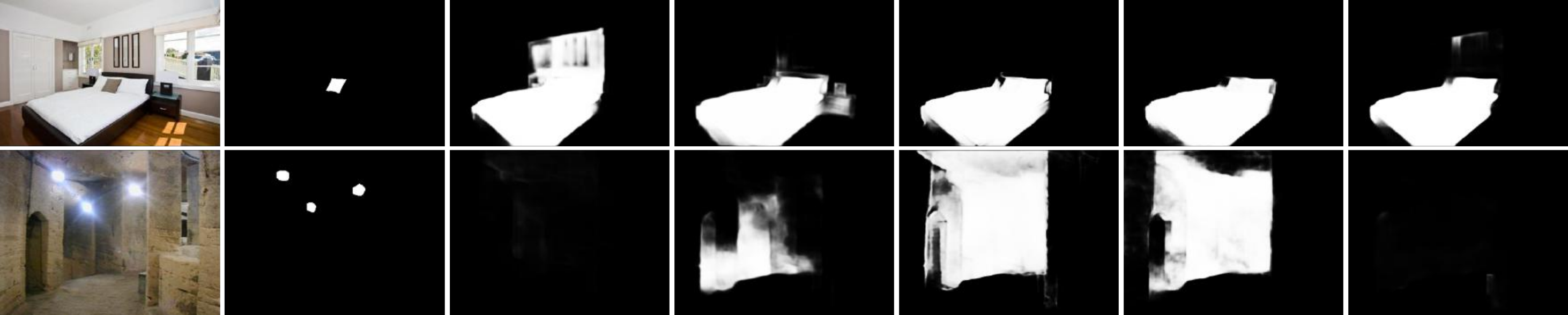}
		\put(2.8,-3){\footnotesize{Image}}
		\put(19.8,-3){\footnotesize{GT}}
		\put(30.8,-3){\footnotesize{Ours-R}}
		\put(45.8,-3){\footnotesize{\cite{zhao2020suppress}}}
		\put(60.5,-3){\footnotesize{\cite{pang2020multi}}}
		\put(74.1,-3){\footnotesize{\cite{zhou2020interactive}}}
		\put(89,-3){\footnotesize{\cite{F3Net}}}
	\end{overpic}
	\caption{Illustration of failure cases. The first and second columns are the input images, and the ground-truth masks. The other columns show the prediction results from our ICON and the baseline methods.}
	\label{fig:failure_case}
\end{figure}
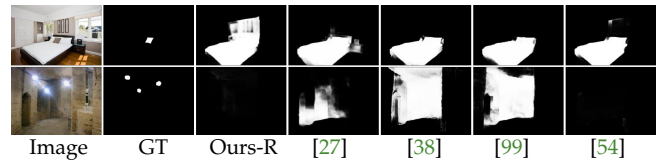

\subsection{Failure Cases}

Although the proposed \ourmodel~method outperforms other SOD algorithms and rarely generates completely incorrect prediction results, there are still \Rev{some failure cases}, as shown in \figref{fig:failure_case}.  
Specifically, in the first row, which shows a tidy room, our method is confused by whether the pillow or the bed and wall is the salient object.
\Rev{Meanwhile, in the second image, the three lamp lights are the salient regions, but our method cannot detect them.}
\Rev{Similarly, other SOTA methods also fail for these samples.}
\Rev{We believe there are several reasons for these failure cases: (1) strong color contrast influencing the model's judgment (\eg, 1$^{st}$ row); (2) lack of sufficient training samples (see the 2$^{nd}$ row) and (3) controversial annotations (\ie, 1$^{st}$ row).
%
}

\subsection{Ablation Study}\label{sec:abl} 
\subsubsection{Effectiveness of Different Components}
To demonstrate the effectiveness of different components in our \ourmodel, we report the quantitative results of several simplified versions of our method.  We start from the encoder-decoder baseline (a UNet-like network with skip connections) and progressively extend it with different modules, including DFA, ICE, and PWV. As shown in \tabref{tab:Baseline}, we first add the DFA (\ie, ID: 2) component upon the Baseline (\ie, ID: 1), which demonstrates an obvious performance improvements. This is reasonable because DFA has the ability to search for objects with 
diverse cues. Then we add the ICE module (\ie, ID: 3), which again shows a substantial performance improvement. \Rev{Finally, as anticipated}, adding all components \Rev{(ID: 4)} to the proposed model achieves the best performance.

\subsubsection{DFA vs. Other Feature Enhancement Methods}
DFA, ASPP~\cite{chen2017deeplab}, Inception~\cite{szegedy2016rethinking}, and PSP~\cite{zhao2017pyramid} are four feature enhancement methods (FEMs), which share some common ideas to stimulate representative feature learning. Differently, our DFA is designed to enhance feature sub-spaces without enlarging the receptive field, which yields more diverse representations.
In \tabref{tab:FEMs}, our DFA with fewer convolutional block clearly outperforms or at least is on par with other FEMs in terms of $S_{m}$, $E_\xi^{m}$, and $F_{\beta}^{w}$. However, DFA also brings some drawbacks. For instance, it generates higher MAE scores when compared with other FEMs. We argue that one possible illustration is that DFA not only brings feature diversity but also some noise. Besides, our experiments~(\ie, ID: 2 vs. ID: 8$\sim$10) reveal that our method after combining three different types of convolution operations can achieve the best score. Meanwhile, our method using only 3xAsyConv generally yields better results than that only using 3xOriConv or 3xAtrConv.

\begin{table*} [t]
	\centering
	\footnotesize
	\renewcommand{\arraystretch}{1.0}
    \setlength\tabcolsep{6.858pt}
	\caption{Ablation analysis of our method when using other routing mechanisms in our PWV. 
	}\label{tab:PWV}
		\begin{tabular}{l|l|cccc|cccc|cccc}
			\toprule
			\multirow{2}{*}{ID} & \multirow{2}{*}{Routing Settings} & \multicolumn{4}{c|}{OMRON~\cite{yang2013saliency}} & \multicolumn{4}{c|}{HKU-IS \cite{li2015visual}} & \multicolumn{4}{c}{DUTS-TE~\cite{wang2017learning}} \\ 
		&&  $S_m$$\uparrow$    & $E_\xi^{m}$$\uparrow$       & $F_\beta^w$$\uparrow$    &   $M$$\downarrow$     &  $S_m$$\uparrow$    & $E_\xi^{m}$$\uparrow$       & $F_\beta^w$$\uparrow$    &   $M$$\downarrow$    &  $S_m$$\uparrow$    & $E_\xi^{m}$$\uparrow$       & $F_\beta^w$$\uparrow$    &   $M$$\downarrow$
			\\ \hline
			\rowcolor{mygray}
            \rowcolor{mygray}
            4 & +DFA+ICE+PWV
            & \textbf{0.844} & \textbf{0.876}  & \textbf{0.761}  & \textbf{0.057}     & 0.920 & \textbf{0.953}  & \textbf{0.902}  & \textbf{0.029}     & \textbf{0.888}  & \textbf{0.924}  & \textbf{0.836}  & \textbf{0.037}	                
            \\
		 14 &
		   \multicolumn{1}{l|}{+DFA+ICE+PWV~(DR)~\cite{sabour2017dynamic}}   &
		   \textbf{0.844}  & 0.868  & 0.757 & 0.058    &  \textbf{0.923} & 0.950  & \textbf{0.902} & 0.030   & \textbf{0.888}  & 0.919  & 0.832 & 0.039
			\\
			15  & \multicolumn{1}{l|}{+DFA+ICE+PWV~(SR)~\cite{hahn2019}} &
			0.837  & 0.862  & 0.745  & 0.060     & 0.912 & 0.843 & 0.895  & 0.030     & 0.881 & 0.903  & 0.831  & 0.042	
			\\ \bottomrule
		\end{tabular}
\end{table*}

\begin{table*} [t!]
    \footnotesize
    \renewcommand{\arraystretch}{1.0}
    \setlength\tabcolsep{8.62pt}
	\centering
    \caption{Ablation analysis of our method when using different loss functions.  
    }\label{tab:loss}
    	\begin{tabular}{l|l|cccc|cccc|cccc}
    		\toprule
    		\multirow{2}{*}{ID} & \multirow{2}{*}{Loss Settings} &\multicolumn{4}{c|}{OMRON~\cite{yang2013saliency}} & \multicolumn{4}{c|}{HKU-IS \cite{li2015visual}} & \multicolumn{4}{c}{DUTS-TE~\cite{wang2017learning}} \\ 
    	&&  $S_m$$\uparrow$    & $E_\xi^{m}$$\uparrow$       & $F_\beta^w$$\uparrow$    &   $M$$\downarrow$     &  $S_m$$\uparrow$    & $E_\xi^{m}$$\uparrow$       & $F_\beta^w$$\uparrow$    &   $M$$\downarrow$    &  $S_m$$\uparrow$    & $E_\xi^{m}$$\uparrow$       & $F_\beta^w$$\uparrow$    &   $M$$\downarrow$
    		\\ \hline
    		\rowcolor{mygray}
    		4 & ICON+$\mathrm{L}_{CPR}$
            & \textbf{0.844} & \textbf{0.876}  & \textbf{0.761}  & \textbf{0.057}     & \textbf{0.920 }& \textbf{0.953}  & \textbf{0.902}  & \textbf{0.029}     & 0.888 & \textbf{0.924}  & \textbf{0.836}  & \textbf{0.037}	                
            \\                
    		16 & \multicolumn{1}{l|}{ICON+$\mathrm{L}_{BCE}$}   &
    		0.840  & 0.866 & 0.757  &  0.060   & 0.918 & 0.950  & 0.899  & 0.031    & \textbf{0.889}  & 0.918  & 0.831  & \textbf{0.037}
    		\\ 
    		\bottomrule
    	\end{tabular}
\end{table*}

\subsection{ICE vs. Attention Methods}
In \tabref{tab:ICE}, we make an additional control group \Rev{(\ie, ID: 3  vs. ID: 11$\sim$13)} to verify the improvement brought by the ICE mechanism. \Rev{Following the same setting (ID: 3), we conduct the experiments to compare ICE with} SE~\cite{hu2020squeeze}, CBAM~\cite{woo2018cbam} and GCT~\cite{yang2020gated}. 
We observe that the alternative method using CBAM achieves an acceptable performance and ranks second among these four methods. However, the other two methods using SE and GCT would lead to a noticeable performance drop. One possible explanation is that our ICE can strengthen the \textit{integrity} of features and highlight potential salient candidates through our designed attention mechanisms.

\subsubsection{Evaluation of Different Routing Algorithms}
To evaluate the performance of EM routing~\cite{hinton2018matrix} \Rev{(ID: 4)}, we also conduct additional experiments \Rev{(see \tabref{tab:PWV})} by replacing it with dynamic routing (DR)~\cite{sabour2017dynamic} and self-routing (SR)~\cite{hahn2019}. We observe that the first alternative method \Rev{(ID: 14)} also achieves reasonable performance, but the second alternative method \Rev{(ID: 15)} yields worse performance, when compared to our method using EM routing. One possible illustration is that SR does not have the routing-by-agreement mechanism, making it incompatible with our PWV scheme.

\subsubsection{Evaluation of Loss Function}
To demonstrate the effectiveness of the $\mathrm{L}_{CPR}$ loss, we conduct another experiment by replacing it as $\mathrm{L}_{BCE}$ in our \ourmodel~architecture. The results reported in \Rev{\tabref{tab:loss}} indicate that, after using the $\mathrm{L}_{CPR}$ loss in the training process, our model can significantly improve the SOD performance across all metrics. Note that combining the IoU and the BCE loss is a common training setting, which has also been used in many recent works~\cite{ qin2019basnet, mao2021transformer}. 

\section{Conclusion}\label{sec5}
We present a novel \textbf{I}ntegrity \textbf{Co}gnition \textbf{N}etwork, called \textbf{\ourmodel}, to detect salient objects from given image scenes. It is based on the observation that mining integral features (at both the micro and macro level) can substantially benefit the salient object detection process. Specifically, in this work, three novel network modules are designed: the diverse feature aggregation module, the integrity channel enhancement module, and the part-whole verification module. By integrating these modules, our \ourmodel~is able to capture diverse features at each feature level and enhance feature channels that highlight the potential integral salient objects, as well as further verify the part-whole agreement between the mined salient object regions. Comprehensive experiments on seven benchmark datasets are conducted. The experimental results demonstrate the contribution of each newly proposed component, as well as the state-of-the-art performance of our \ourmodel.  

\ifCLASSOPTIONcompsoc
  \section*{Acknowledgments}
\else
  \section*{Acknowledgment}
\fi
The authors would like to thank the anonymous reviewers and editor for their helpful comments on this manuscript. And we thank Jing Zhang for sharing codes of their work.
This work is partially funded by the National Key R\&D Program of China under Grant 2021B0101200001; the National Natural Science Foundation of China under Grant  U21B2048 and 61929104, and Zhejiang Lab (No.2019KD0AD01/010).

{
\small
\bibliographystyle{IEEEtran}
\bibliography{icon}

\begin{thebibliography}{100}
\providecommand{\url}[1]{#1}
\csname url@samestyle\endcsname
\providecommand{\newblock}{\relax}
\providecommand{\bibinfo}[2]{#2}
\providecommand{\BIBentrySTDinterwordspacing}{\spaceskip=0pt\relax}
\providecommand{\BIBentryALTinterwordstretchfactor}{4}
\providecommand{\BIBentryALTinterwordspacing}{\spaceskip=\fontdimen2\font plus
\BIBentryALTinterwordstretchfactor\fontdimen3\font minus
  \fontdimen4\font\relax}
\providecommand{\BIBforeignlanguage}[2]{{%
\expandafter\ifx\csname l@#1\endcsname\relax
\typeout{** WARNING: IEEEtran.bst: No hyphenation pattern has been}%
\typeout{** loaded for the language `#1'. Using the pattern for}%
\typeout{** the default language instead.}%
\else
\language=\csname l@#1\endcsname
\fi
#2}}
\providecommand{\BIBdecl}{\relax}
\BIBdecl

\bibitem{borji2015salient}
A.~Borji, M.-M. Cheng, H.~Jiang, and J.~Li, ``{Salient object detection: A
  benchmark},'' \emph{{IEEE Trans. Image Process.}}, vol.~24, no.~12, pp.
  5706--5722, 2015.

\bibitem{fan2019rethinking}
D.-P. Fan, Z.~Lin, Z.~Zhang, M.~Zhu, and M.-M. Cheng, ``Rethinking rgb-d
  salient object detection: Models, data sets, and large-scale benchmarks,''
  \emph{IEEE Trans. Neural Netw. Learn. Syst.}, vol.~32, no.~5, pp. 2075--2089,
  2021.

\bibitem{wang2019salient1}
W.~Wang, Q.~Lai, H.~Fu, J.~Shen, and H.~Ling, ``{Salient object detection in
  the deep learning era: An in-depth survey},'' \emph{{IEEE Trans. Pattern
  Anal. Mach. Intell.}}, 2021.

\bibitem{zhang2019IJCV}
D.~Zhang, J.~Han, L.~Zhao, and D.~Meng, ``Leveraging prior-knowledge for weakly
  supervised object detection under a collaborative self-paced curriculum
  learning framework,'' \emph{{Int. J. Comput. Vis.}}, vol. 127, no.~4, pp.
  363--380, 2019.

\bibitem{liu2013model}
G.~Liu and D.~Fan, ``A model of visual attention for natural image retrieval,''
  in \emph{Int. Conf. Inf. Sci. Cloud Comput. Companion}, 2013, pp. 728--733.

\bibitem{deng2020re}
D.-P. Fan, T.~Li, Z.~Lin, G.-P. Ji, D.~Zhang, M.-M. Cheng, H.~Fu, and J.~Shen,
  ``Re-thinking co-salient object detection,'' \emph{{IEEE Trans. Pattern Anal.
  Mach. Intell.}}, 2022.

\bibitem{Zhuge2021KaleidoBERT}
M.~Zhuge, D.~Gao, D.-P. Fan, L.~Jin, B.~Chen, H.~Zhou, M.~Qiu, and L.~Shao,
  ``Kaleido-bert: Vision-language pre-training on fashion domain,'' in
  \emph{{IEEE Conf. Comput. Vis. Pattern Recog.}}, 2021, pp. 12\,647--12\,657.

\bibitem{qin2020u2}
X.~Qin, Z.~Zhang, C.~Huang, M.~Dehghan, O.~R. Zaiane, and M.~Jagersand,
  ``U2-net: Going deeper with nested u-structure for salient object
  detection,'' \emph{Pattern Recognition}, vol. 106, p. 107404, 2020.

\bibitem{hoyer2019grid}
L.~Hoyer, M.~Munoz, P.~Katiyar, A.~Khoreva, and V.~Fischer, ``Grid saliency for
  context explanations of semantic segmentation,'' in \emph{{Adv. Neural
  Inform. Process. Syst.}}, 2019, pp. 6462--6473.

\bibitem{wei2016stc}
Y.~Wei, X.~Liang, Y.~Chen, X.~Shen, M.-M. Cheng, J.~Feng, Y.~Zhao, and S.~Yan,
  ``Stc: A simple to complex framework for weakly-supervised semantic
  segmentation,'' \emph{{IEEE Trans. Pattern Anal. Mach. Intell.}}, vol.~39,
  no.~11, pp. 2314--2320, 2016.

\bibitem{zeng2019joint}
Y.~Zeng, Y.~Zhuge, H.~Lu, and L.~Zhang, ``Joint learning of saliency detection
  and weakly supervised semantic segmentation,'' in \emph{{IEEE Int. Conf.
  Comput. Vis.}}, 2019, pp. 7223--7233.

\bibitem{wang2017saliency}
W.~Wang, J.~Shen, R.~Yang, and F.~Porikli, ``{Saliency-aware video object
  segmentation},'' \emph{{IEEE Trans. Pattern Anal. Mach. Intell.}}, vol.~40,
  no.~1, pp. 20--33, 2017.

\bibitem{cheng2014global}
M.-M. Cheng, N.~J. Mitra, X.~Huang, P.~H. Torr, and S.-M. Hu, ``{Global
  contrast based salient region detection},'' \emph{{IEEE Trans. Pattern Anal.
  Mach. Intell.}}, vol.~37, no.~3, pp. 569--582, 2014.

\bibitem{itti1998model}
L.~Itti, C.~Koch, and E.~Niebur, ``A model of saliency-based visual attention
  for rapid scene analysis,'' \emph{{IEEE Trans. Pattern Anal. Mach. Intell.}},
  vol.~20, no.~11, pp. 1254--1259, 1998.

\bibitem{yang2013saliency}
C.~Yang, L.~Zhang, H.~Lu, X.~Ruan, and M.-H. Yang, ``{Saliency detection via
  graph-based manifold ranking},'' in \emph{{IEEE Conf. Comput. Vis. Pattern
  Recog.}}, 2013, pp. 3166--3173.

\bibitem{zhu2014saliency}
W.~Zhu, S.~Liang, Y.~Wei, and J.~Sun, ``{Saliency optimization from robust
  background detection},'' in \emph{{IEEE Conf. Comput. Vis. Pattern Recog.}},
  2014, pp. 2814--2821.

\bibitem{jiang2013salient}
H.~Jiang, J.~Wang, Z.~Yuan, Y.~Wu, N.~Zheng, and S.~Li, ``{Salient object
  detection: A discriminative regional feature integration approach},'' in
  \emph{{IEEE Conf. Comput. Vis. Pattern Recog.}}, 2013, pp. 2083--2090.

\bibitem{fan2018SOC}
D.-P. Fan, M.-M. Cheng, J.-J. Liu, S.-H. Gao, Q.~Hou, and A.~Borji, ``Salient
  objects in clutter: Bringing salient object detection to the foreground,'' in
  \emph{{Eur. Conf. Comput. Vis.}}, 2018, pp. 186--202.

\bibitem{borji2014salient}
A.~Borji, M.-M. Cheng, Q.~Hou, H.~Jiang, and J.~Li, ``Salient object detection:
  A survey,'' \emph{{Comput. Vis. Media}}, pp. 1--34, 2014.

\bibitem{han2018advanced}
J.~Han, D.~Zhang, G.~Cheng, N.~Liu, and D.~Xu, ``Advanced deep-learning
  techniques for salient and category-specific object detection: a survey,''
  \emph{{IEEE Signal Process. Mag.}}, vol.~35, no.~1, pp. 84--100, 2018.

\bibitem{cheng2021boundary}
B.~Cheng, R.~Girshick, P.~Doll{\'a}r, A.~C. Berg, and A.~Kirillov, ``Boundary
  iou: Improving object-centric image segmentation evaluation,'' in \emph{{IEEE
  Conf. Comput. Vis. Pattern Recog.}}, 2021, pp. 15\,334--15\,342.

\bibitem{zhang2017amulet}
P.~Zhang, D.~Wang, H.~Lu, H.~Wang, and X.~Ruan, ``Amulet: Aggregating
  multi-level convolutional features for salient object detection,'' in
  \emph{{IEEE Int. Conf. Comput. Vis.}}, 2017, pp. 202--211.

\bibitem{8100181}
Z.~{Luo}, A.~{Mishra}, A.~{Achkar}, J.~{Eichel}, S.~{Li}, and P.~{Jodoin},
  ``Non-local deep features for salient object detection,'' in \emph{{IEEE
  Conf. Comput. Vis. Pattern Recog.}}, 2017, pp. 6593--6601.

\bibitem{zhao2019pyramid}
T.~Zhao and X.~Wu, ``Pyramid feature attention network for saliency
  detection,'' in \emph{{IEEE Conf. Comput. Vis. Pattern Recog.}}, 2019, pp.
  3085--3094.

\bibitem{liu2020picanet}
{N. Liu, J. Han, and M.-H. Yang}, ``Picanet: Pixel-wise contextual attention
  learning for accurate saliency detection,'' \emph{{IEEE Trans. Image
  Process.}}, vol.~29, pp. 6438--6451, 2020.

\bibitem{wang2019iterative}
W.~Wang, J.~Shen, M.-M. Cheng, and L.~Shao, ``An iterative and cooperative
  top-down and bottom-up inference network for salient object detection,'' in
  \emph{{IEEE Conf. Comput. Vis. Pattern Recog.}}, 2019, pp. 5968--5977.

\bibitem{zhao2020suppress}
X.~Zhao, Y.~Pang, L.~Zhang, H.~Lu, and L.~Zhang, ``Suppress and balance: A
  simple gated network for salient object detection,'' in \emph{{Eur. Conf.
  Comput. Vis.}}, 2020, pp. 35--51.

\bibitem{liu2019simple}
J.-J. Liu, Q.~Hou, M.-M. Cheng, J.~Feng, and J.~Jiang, ``{A simple
  pooling-based design for real-time salient object detection},'' in
  \emph{{IEEE Conf. Comput. Vis. Pattern Recog.}}, 2019, pp. 3917--3926.

\bibitem{wei2020label}
J.~Wei, S.~Wang, Z.~Wu, C.~Su, Q.~Huang, and Q.~Tian, ``Label decoupling
  framework for salient object detection,'' in \emph{{IEEE Conf. Comput. Vis.
  Pattern Recog.}}, 2020, pp. 13\,025--13\,034.

\bibitem{wu2019mutual}
R.~Wu, M.~Feng, W.~Guan, D.~Wang, H.~Lu, and E.~Ding, ``A mutual learning
  method for salient object detection with intertwined multi-supervision,'' in
  \emph{{IEEE Conf. Comput. Vis. Pattern Recog.}}, 2019, pp. 8150--8159.

\bibitem{amirul2018revisiting}
M.~Amirul~Islam, M.~Kalash, and N.~D. Bruce, ``Revisiting salient object
  detection: Simultaneous detection, ranking, and subitizing of multiple
  salient objects,'' in \emph{{IEEE Conf. Comput. Vis. Pattern Recog.}}, 2018,
  pp. 7142--7150.

\bibitem{8237382}
S.~{He}, J.~{Jiao}, X.~{Zhang}, G.~{Han}, and R.~W.~H. {Lau}, ``Delving into
  salient object subitizing and detection,'' in \emph{{IEEE Int. Conf. Comput.
  Vis.}}, 2017, pp. 1059--1067.

\bibitem{hinton2018matrix}
G.~E. Hinton, S.~Sabour, and N.~Frosst, ``Matrix capsules with em routing,'' in
  \emph{{Int. Conf. Learn. Represent.}}, 2018.

\bibitem{xie2017holistically}
S.~Xie and Z.~Tu, ``Holistically-nested edge detection,'' \emph{IJCV}, vol.
  125, no. 1-3, pp. 3--18, 2017.

\bibitem{hou2019deeply}
Q.~Hou, M.-M. Cheng, X.~Hu, A.~Borji, Z.~Tu, and P.~H. Torr, ``Deeply
  supervised salient object detection with short connections,'' \emph{PAMI},
  vol.~41, no.~4, pp. 815--828, 2019.

\bibitem{hu18recurrently}
X.~Hu, L.~Zhu, J.~Qin, C.-W. Fu, and P.-A. Heng, ``Recurrently aggregating deep
  features for salient object detection,'' in \emph{{AAAI Conf. Art. Intell.}},
  2018, pp. 6943--6950.

\bibitem{zhao2019optimizing}
K.~Zhao, S.~Gao, W.~Wang, and M.-M. Cheng, ``{Optimizing the f-measure for
  threshold-free salient object detection},'' in \emph{{IEEE Int. Conf. Comput.
  Vis.}}, 2019, pp. 8849--8857.

\bibitem{pang2020multi}
Y.~Pang, X.~Zhao, L.~Zhang, and H.~Lu, ``Multi-scale interactive network for
  salient object detection,'' in \emph{{IEEE Conf. Comput. Vis. Pattern
  Recog.}}, 2020, pp. 9413--9422.

\bibitem{GateNet}
X.~Zhao, Y.~Pang, L.~Zhang, H.~Lu, and L.~Zhang, ``Suppress and balance: A
  simple gated network for salient object detection,'' in \emph{{Eur. Conf.
  Comput. Vis.}}, 2020, pp. 35--51.

\bibitem{liu2021rethinking}
J.-J. Liu, Z.-A. Liu, P.~Peng, and M.-M. Cheng, ``Rethinking the u-shape
  structure for salient object detection,'' \emph{TIP}, vol.~30, pp.
  9030--9042, 2021.

\bibitem{ma2003contrast}
Y.-F. Ma and H.-J. Zhang, ``Contrast-based image attention analysis by using
  fuzzy growing,'' in \emph{{ACM Int. Conf. Multimedia}}, 2003, pp. 374--381.

\bibitem{harel2006graph}
J.~Harel, C.~Koch, and P.~Perona, ``Graph-based visual saliency,'' in
  \emph{{Adv. Neural Inform. Process. Syst.}}, 2006.

\bibitem{hu2017deep}
P.~Hu, B.~Shuai, J.~Liu, and G.~Wang, ``Deep level sets for salient object
  detection,'' in \emph{{IEEE Conf. Comput. Vis. Pattern Recog.}}, 2017, pp.
  2300--2309.

\bibitem{he2015supercnn}
S.~He, R.~W. Lau, W.~Liu, Z.~Huang, and Q.~Yang, ``{Supercnn: A superpixelwise
  convolutional neural network for salient object detection},'' \emph{{Int. J.
  Comput. Vis.}}, vol. 115, no.~3, pp. 330--344, 2015.

\bibitem{li2016deep}
G.~Li and Y.~Yu, ``Deep contrast learning for salient object detection,'' in
  \emph{{IEEE Conf. Comput. Vis. Pattern Recog.}}, 2016, pp. 478--487.

\bibitem{liu2018picanet}
N.~Liu, J.~Han, and M.-H. Yang, ``{PiCANet: Learning pixel-wise contextual
  attention for saliency detection},'' in \emph{{IEEE Conf. Comput. Vis.
  Pattern Recog.}}, 2018, pp. 3089--3098.

\bibitem{lafferty2001conditional}
J.~D. Lafferty, A.~McCallum, and F.~C. Pereira, ``Conditional random fields:
  Probabilistic models for segmenting and labeling sequence data,'' in
  \emph{ICML}, 2001.

\bibitem{luo2017non}
Z.~Luo, A.~Mishra, A.~Achkar, J.~Eichel, S.~Li, and P.-M. Jodoin, ``Non-local
  deep features for salient object detection,'' in \emph{{IEEE Conf. Comput.
  Vis. Pattern Recog.}}, 2017, pp. 6593--6601.

\bibitem{li2018contour}
X.~Li, F.~Yang, H.~Cheng, W.~Liu, and D.~Shen, ``{Contour knowledge transfer
  for salient object detection},'' in \emph{{Eur. Conf. Comput. Vis.}}, 2018,
  pp. 355--370.

\bibitem{su2019selectivity}
J.~Su, J.~Li, Y.~Zhang, C.~Xia, and Y.~Tian, ``Selectivity or invariance:
  Boundary-aware salient object detection,'' in \emph{{IEEE Int. Conf. Comput.
  Vis.}}, 2019, pp. 3799--3808.

\bibitem{zhao2019egnet}
J.-X. Zhao, J.-J. Liu, D.-P. Fan, Y.~Cao, J.~Yang, and M.-M. Cheng, ``{EGNet:
  Edge guidance network for salient object detection},'' in \emph{{IEEE Int.
  Conf. Comput. Vis.}}, 2019, pp. 8779--8788.

\bibitem{qin2019basnet}
X.~Qin, Z.~Zhang, C.~Huang, C.~Gao, M.~Dehghan, and M.~Jagersand, ``{BASNet:
  Boundary-aware salient object detection},'' in \emph{{IEEE Conf. Comput. Vis.
  Pattern Recog.}}, 2019, pp. 7479--7489.

\bibitem{zeng2019towards}
Y.~Zeng, P.~Zhang, J.~Zhang, Z.~Lin, and H.~Lu, ``Towards high-resolution
  salient object detection,'' in \emph{{IEEE Int. Conf. Comput. Vis.}}, 2019,
  pp. 7234--7243.

\bibitem{F3Net}
J.~Wei, S.~Wang, and Q.~Huang, ``F$^3$net: Fusion, feedback and focus for
  salient object detection,'' in \emph{{AAAI Conf. Art. Intell.}}, vol.~34,
  no.~07, 2020, pp. 12\,321--12\,328.

\bibitem{wu2019stacked}
Z.~Wu, L.~Su, and Q.~Huang, ``{Stacked Cross Refinement Network for Edge-Aware
  Salient Object Detection},'' in \emph{{IEEE Int. Conf. Comput. Vis.}}, 2019,
  pp. 7264--7273.

\bibitem{liu2021visual}
N.~Liu, N.~Zhang, K.~Wan, L.~Shao, and J.~Han, ``Visual saliency transformer,''
  in \emph{{IEEE Int. Conf. Comput. Vis.}}, 2021.

\bibitem{he2016deep}
K.~He, X.~Zhang, S.~Ren, and J.~Sun, ``{Deep residual learning for image
  recognition},'' in \emph{{IEEE Conf. Comput. Vis. Pattern Recog.}}, 2016, pp.
  770--778.

\bibitem{wu2019cascaded}
Z.~Wu, L.~Su, and Q.~Huang, ``{Cascaded partial decoder for fast and accurate
  salient object detection},'' in \emph{{IEEE Conf. Comput. Vis. Pattern
  Recog.}}, 2019, pp. 3907--3916.

\bibitem{liu2019employing}
Y.~Liu, Q.~Zhang, D.~Zhang, and J.~Han, ``Employing deep part-object
  relationships for salient object detection,'' in \emph{{IEEE Int. Conf.
  Comput. Vis.}}, 2019, pp. 1232--1241.

\bibitem{GCPANet}
{Zuyao Chen and Qianqian Xu and Runmin Cong and Qingming Huang}, ``{Global
  Context-Aware Progressive Aggregation Network for Salient Object
  Detection},'' in \emph{{AAAI Conf. Art. Intell.}}, vol.~34, no.~07, 2020, pp.
  10\,599--10\,606.

\bibitem{wu2020deeper}
Z.~Wu, S.~Li, C.~Chen, A.~Hao, and H.~Qin, ``A deeper look at image salient
  object detection: Bi-stream network with a small training dataset,''
  \emph{{IEEE Trans. Multimedia}}, 2020.

\bibitem{mao2021transformer}
Y.~Mao, J.~Zhang, Z.~Wan, Y.~Dai, A.~Li, Y.~Lv, X.~Tian, D.-P. Fan, and
  N.~Barnes, ``Generative transformer for accurate and reliable salient object
  detection,'' \emph{arXiv preprint arXiv:2104.10127}, 2021.

\bibitem{chen2017rethinking}
L.-C. Chen, G.~Papandreou, F.~Schroff, and H.~Adam, ``Rethinking atrous
  convolution for semantic image segmentation,'' \emph{arXiv preprint
  arXiv:1706.05587}, 2017.

\bibitem{szegedy2016rethinking}
C.~Szegedy, V.~Vanhoucke, S.~Ioffe, J.~Shlens, and Z.~Wojna, ``Rethinking the
  inception architecture for computer vision,'' in \emph{{IEEE Conf. Comput.
  Vis. Pattern Recog.}}, 2016, pp. 2818--2826.

\bibitem{zhao2017pyramid}
H.~Zhao, J.~Shi, X.~Qi, X.~Wang, and J.~Jia, ``Pyramid scene parsing network,''
  in \emph{{IEEE Conf. Comput. Vis. Pattern Recog.}}, 2017, pp. 2881--2890.

\bibitem{ding2019acnet}
X.~Ding, Y.~Guo, G.~Ding, and J.~Han, ``{ACNet: Strengthening the kernel
  skeletons for powerful cnn via asymmetric convolution blocks},'' in
  \emph{{IEEE Int. Conf. Comput. Vis.}}, 2019, pp. 1911--1920.

\bibitem{chen2017deeplab}
L.-C. Chen, G.~Papandreou, I.~Kokkinos, K.~Murphy, and A.~L. Yuille,
  ``{Deeplab: Semantic image segmentation with deep convolutional nets, atrous
  convolution, and fully connected crfs},'' \emph{{IEEE Trans. Pattern Anal.
  Mach. Intell.}}, vol.~40, no.~4, pp. 834--848, 2017.

\bibitem{hu2020squeeze}
J.~Hu, L.~Shen, S.~Albanie, G.~Sun, and E.~Wu, ``Squeeze-and-excitation
  networks,'' \emph{PAMI}, vol.~42, no.~08, pp. 2011--2023, 2020.

\bibitem{woo2018cbam}
S.~Woo, J.~Park, J.-Y. Lee, and I.~So~Kweon, ``Cbam: Convolutional block
  attention module,'' in \emph{{Eur. Conf. Comput. Vis.}}, 2018, pp. 3--19.

\bibitem{cao2019gcnet}
Y.~Cao, J.~Xu, S.~Lin, F.~Wei, and H.~Hu, ``Gcnet: Non-local networks meet
  squeeze-excitation networks and beyond,'' in \emph{{IEEE Int. Conf. Comput.
  Vis. Worksh.}}, 2019, pp. 0--0.

\bibitem{yang2020gated}
Z.~Yang, L.~Zhu, Y.~Wu, and Y.~Yang, ``Gated channel transformation for visual
  recognition,'' in \emph{{IEEE Conf. Comput. Vis. Pattern Recog.}}, 2020, pp.
  11\,794--11\,803.

\bibitem{sabour2017dynamic}
S.~Sabour, N.~Frosst, and G.~E. Hinton, ``Dynamic routing between capsules,''
  in \emph{{Adv. Neural Inform. Process. Syst.}}, 2017, pp. 3856--3866.

\bibitem{lalonde2018capsules}
R.~LaLonde and U.~Bagci, ``Capsules for object segmentation,'' in
  \emph{International conference on Medical Imaging with Deep Learning}, 2018.

\bibitem{mattyus2017deeproadmapper}
G.~M{\'a}ttyus, W.~Luo, and R.~Urtasun, ``Deeproadmapper: Extracting road
  topology from aerial images,'' in \emph{{IEEE Int. Conf. Comput. Vis.}},
  2017, pp. 3438--3446.

\bibitem{wang2017learning}
L.~Wang, H.~Lu, Y.~Wang, M.~Feng, D.~Wang, B.~Yin, and X.~Ruan, ``{Learning to
  detect salient objects with image-level supervision},'' in \emph{{IEEE Conf.
  Comput. Vis. Pattern Recog.}}, 2017, pp. 136--145.

\bibitem{yan2013hierarchical}
Q.~Yan, L.~Xu, J.~Shi, and J.~Jia, ``{Hierarchical saliency detection},'' in
  \emph{{IEEE Conf. Comput. Vis. Pattern Recog.}}, 2013, pp. 1155--1162.

\bibitem{li2015visual}
G.~Li and Y.~Yu, ``{Visual saliency based on multiscale deep features},'' in
  \emph{{IEEE Conf. Comput. Vis. Pattern Recog.}}, 2015, pp. 5455--5463.

\bibitem{li2014secrets}
Y.~Li, X.~Hou, C.~Koch, J.~M. Rehg, and A.~L. Yuille, ``{The secrets of salient
  object segmentation},'' in \emph{{IEEE Conf. Comput. Vis. Pattern Recog.}},
  2014, pp. 280--287.

\bibitem{movahedi2010design}
V.~Movahedi and J.~H. Elder, ``{Design and perceptual validation of performance
  measures for salient object segmentation},'' in \emph{{IEEE Conf. Comput.
  Vis. Pattern Recog. Worksh.}}, 2010, pp. 49--56.

\bibitem{simonyan2014very}
K.~Simonyan and A.~Zisserman, ``Very deep convolutional networks for
  large-scale image recognition,'' in \emph{{Int. Conf. Learn. Represent.}},
  2015.

\bibitem{wang2021pvtv2}
{W. Wang, E. Xie, X. Li, D.-P. Fan, K. Song, D. Liang, T. Lu, P. Luo, and L.
  Shao}, ``Pvtv2: Improved baselines with pyramid vision transformer,''
  \emph{{Comput. Vis. Media}}, vol.~8, no.~3, pp. 1--10, 2022.

\bibitem{liu2021swin}
Z.~Liu, Y.~Lin, Y.~Cao, H.~Hu, Y.~Wei, Z.~Zhang, S.~Lin, and B.~Guo, ``Swin
  transformer: Hierarchical vision transformer using shifted windows,'' in
  \emph{{IEEE Int. Conf. Comput. Vis.}}, 2021.

\bibitem{chen2021cyclemlp}
S.~Chen, E.~Xie, C.~Ge, D.~Liang, and P.~Luo, ``Cyclemlp: A mlp-like
  architecture for dense prediction,'' in \emph{{Int. Conf. Learn.
  Represent.}}, 2022.

\bibitem{wang2021pyramid}
W.~Wang, E.~Xie, X.~Li, D.-P. Fan, K.~Song, D.~Liang, T.~Lu, P.~Luo, and
  L.~Shao, ``Pyramid vision transformer: A versatile backbone for dense
  prediction without convolutions,'' in \emph{{IEEE Int. Conf. Comput. Vis.}},
  2021.

\bibitem{he2015delving}
K.~He, X.~Zhang, S.~Ren, and J.~Sun, ``Delving deep into rectifiers: Surpassing
  human-level performance on imagenet classification,'' in \emph{{IEEE Int.
  Conf. Comput. Vis.}}, 2015, pp. 1026--1034.

\bibitem{bottou2012stochastic}
L.~Bottou, ``Stochastic gradient descent tricks,'' in \emph{Neural networks:
  Tricks of the trade}, 2012, pp. 421--436.

\bibitem{margolin2014evaluate}
R.~Margolin, L.~Zelnik-Manor, and A.~Tal, ``How to evaluate foreground maps?''
  in \emph{{IEEE Conf. Comput. Vis. Pattern Recog.}}, 2014, pp. 248--255.

\bibitem{zhang2020multistage}
L.~Zhang, J.~Wu, T.~Wang, A.~Borji, G.~Wei, and H.~Lu, ``A multistage
  refinement network for salient object detection,'' \emph{TIP}, vol.~29, pp.
  3534--3545, 2020.

\bibitem{wang2018salient2}
L.~Wang, L.~Wang, H.~Lu, P.~Zhang, and X.~Ruan, ``Salient object detection with
  recurrent fully convolutional networks,'' \emph{{IEEE Trans. Pattern Anal.
  Mach. Intell.}}, vol.~41, no.~7, pp. 1734--1746, 2018.

\bibitem{piao2019depth}
Y.~Piao, W.~Ji, J.~Li, M.~Zhang, and H.~Lu, ``{Depth-Induced Multi-Scale
  Recurrent Attention Network for Saliency Detection},'' in \emph{{IEEE Int.
  Conf. Comput. Vis.}}, 2019, pp. 7254--7263.

\bibitem{li2017instance}
G.~Li, Y.~Xie, L.~Lin, and Y.~Yu, ``Instance-level salient object
  segmentation,'' in \emph{{IEEE Conf. Comput. Vis. Pattern Recog.}}, 2017, pp.
  2386--2395.

\bibitem{achanta2009frequency}
R.~Achanta, S.~Hemami, F.~Estrada, and S.~Susstrunk, ``Frequency-tuned salient
  region detection,'' in \emph{{IEEE Conf. Comput. Vis. Pattern Recog.}}, 2009,
  pp. 1597--1604.

\bibitem{cheng2021structure}
M.-M. Cheng and D.-P. Fan, ``Structure-measure: A new way to evaluate
  foreground maps,'' \emph{IJCV}, vol. 129, no.~9, pp. 2622--2638, 2021.

\bibitem{fan2018enhanced}
D.-P. Fan, G.-P. Ji, X.~Qin, and M.-M. Cheng, ``Cognitive vision inspired
  object segmentation metric and loss function,'' \emph{SCIENTIA SINICA
  Informationis}, 2021.

\bibitem{tian2020conditional}
Z.~Tian, C.~Shen, and H.~Chen, ``Conditional convolutions for instance
  segmentation,'' in \emph{ECCV}.\hskip 1em plus 0.5em minus 0.4em\relax
  Springer, 2020, pp. 282--298.

\bibitem{kirillov2020pointrend}
A.~Kirillov, Y.~Wu, K.~He, and R.~Girshick, ``Pointrend: Image segmentation as
  rendering,'' in \emph{CVPR}, 2020, pp. 9799--9808.

\bibitem{chen2018reverse}
S.~Chen, X.~Tan, B.~Wang, and X.~Hu, ``{Reverse attention for salient object
  detection},'' in \emph{{Eur. Conf. Comput. Vis.}}, 2018, pp. 234--250.

\bibitem{feng2019attentive}
M.~Feng, H.~Lu, and E.~Ding, ``{Attentive feedback network for boundary-aware
  salient object detection},'' in \emph{{IEEE Conf. Comput. Vis. Pattern
  Recog.}}, 2019, pp. 1623--1632.

\bibitem{zhou2020interactive}
H.~Zhou, X.~Xie, J.-H. Lai, Z.~Chen, and L.~Yang, ``Interactive two-stream
  decoder for accurate and fast saliency detection,'' in \emph{{IEEE Conf.
  Comput. Vis. Pattern Recog.}}, 2020, pp. 9141--9150.

\bibitem{DBLP:conf/iccv/WangBZZL17}
T.~Wang, A.~Borji, L.~Zhang, P.~Zhang, and H.~Lu, ``A stagewise refinement
  model for detecting salient objects in images,'' in \emph{{IEEE Int. Conf.
  Comput. Vis.}}, 2017, pp. 4039--4048.

\bibitem{deng2018r3net}
Z.~Deng, X.~Hu, L.~Zhu, X.~Xu, J.~Qin, G.~Han, and P.-A. Heng, ``{R3Net:
  Recurrent residual refinement network for saliency detection},'' in
  \emph{{Int. Joint Conf. Artif. Intell.}}, 2018, pp. 684--690.

\bibitem{zhang2018bi}
L.~Zhang, J.~Dai, H.~Lu, Y.~He, and G.~Wang, ``A bi-directional message passing
  model for salient object detection,'' in \emph{{IEEE Conf. Comput. Vis.
  Pattern Recog.}}, 2018, pp. 1741--1750.

\bibitem{wang2018detect}
T.~Wang, L.~Zhang, S.~Wang, H.~Lu, G.~Yang, X.~Ruan, and A.~Borji, ``Detect
  globally, refine locally: A novel approach to saliency detection,'' in
  \emph{{IEEE Conf. Comput. Vis. Pattern Recog.}}, 2018, pp. 3127--3135.

\bibitem{chen2020reverse}
S.~Chen, X.~Tan, B.~Wang, H.~Lu, X.~Hu, and Y.~Fu, ``Reverse attention-based
  residual network for salient object detection,'' \emph{{IEEE Trans. Image
  Process.}}, vol.~29, pp. 3763--3776, 2020.

\bibitem{Fan2021SOC}
D.-P. Fan, J.~Zhang, G.~Xu, M.-M. Cheng, and L.~Shao, ``Salient objects in
  clutter,'' \emph{{IEEE Trans. Pattern Anal. Mach. Intell.}}, 2022.

\bibitem{hahn2019}
T.~Hahn, M.~Pyeon, and G.~Kim, ``Self-routing capsule networks,'' in
  \emph{{Adv. Neural Inform. Process. Syst.}}, 2019, pp. 7658--7667.

\end{thebibliography}
}

\clearpage

\vspace{-30pt}

\vfill

\end{document}